\documentclass[10pt,twocolumn,letterpaper]{article}
\usepackage{cvpr}
\usepackage{times}
\usepackage{amsmath}
\usepackage{amssymb}
\usepackage{amssymb}
\usepackage{booktabs}
\usepackage{physics}
\usepackage{caption}
\usepackage{enumitem}
\usepackage{fontawesome}
\usepackage[dvipsnames]{xcolor}
\usepackage{graphicx}
\usepackage{wrapfig}
\usepackage{textcomp}%
\usepackage{multirow}
\usepackage[para]{footmisc}

\usepackage[pagebackref,breaklinks,colorlinks,citecolor=cvprblue]{hyperref}
\usepackage[capitalize]{cleveref}

\definecolor{cvprblue}{rgb}{0.21,0.49,0.74}

\setlist[itemize]{leftmargin=1em}
\setlist[enumerate]{leftmargin=1em}

\Crefname{section}{Sec.}{Secs.}
\Crefname{equation}{Eq.}{Eqs.}
\Crefname{figure}{Fig.}{Figs.}
\Crefname{tabular}{Tab.}{Tabs.}

\newcommand{\method}{Farm3D\xspace}
\newcommand{\magicpony}{MagicPony\xspace}

\makeatletter
\renewcommand{\paragraph}{%
  \@startsection{paragraph}{4}%
  {\z@}{0.25em}{-1em}%
  {\normalfont\normalsize\bfseries}%
}
\makeatother

\def\paperID{164} %
\def\confName{3DV\xspace}
\def\confYear{2024\xspace}

\title{\method: Learning Articulated 3D Animals by Distilling 2D Diffusion}

\author{Tomas Jakab\thanks{Equal contribution.}
\quad
Ruining Li$^*$
\quad
Shangzhe Wu
\quad
Christian Rupprecht
\quad
Andrea Vedaldi \\[0.3em]
Visual Geometry Group, University of Oxford\\
{\tt\small \{tomj, ruining, szwu, chrisr, vedaldi\}@robots.ox.ac.uk}\\[0.1em]
}

\begin{document}
\twocolumn[\maketitle\vspace{-3em}\begin{center}
    \includegraphics[trim={10px 0px 20px 0px}, clip, width=1\linewidth]{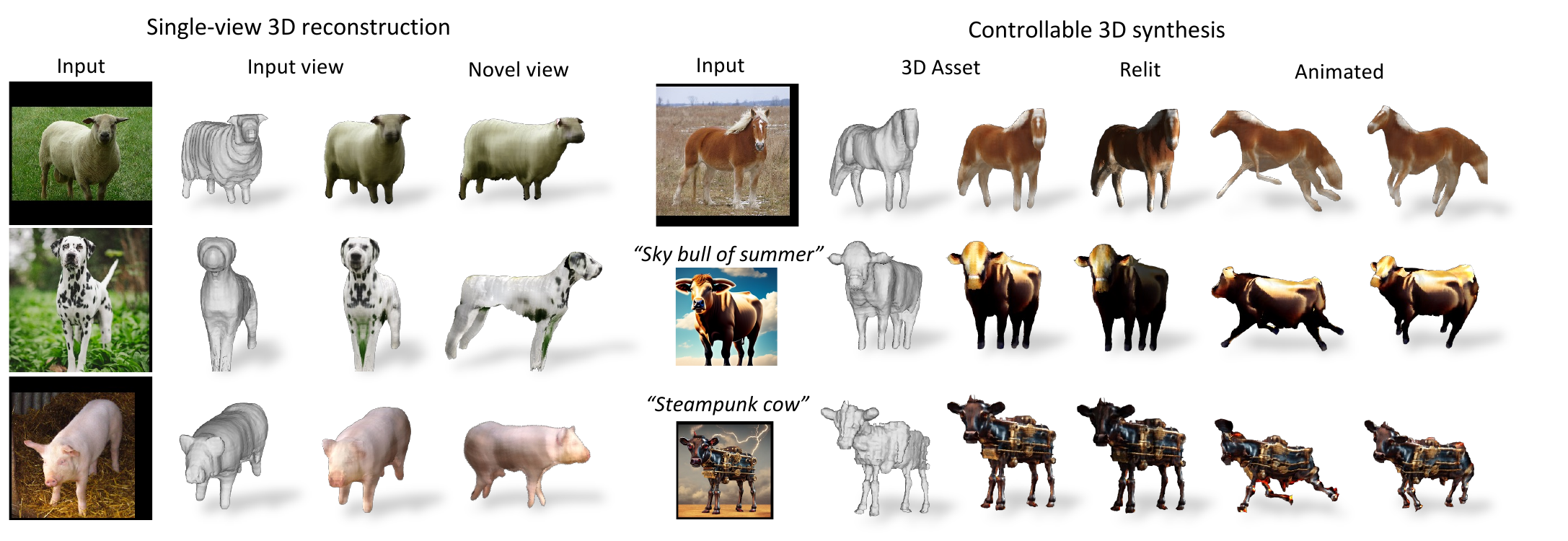}
\end{center}\vspace{-1.5em}
\captionof{figure}{
\textbf{Learning to Reconstruct 3D Animal Categories Purely from Synthetic Images.}
Our method learns to reconstruct articulated and textured animals from single images, using only virtual supervision from an off-the-shelf diffusion-based 2D image generator, without the curation of any real training images.
The method generalizes to a wide range of animal categories, such as cows, horses, sheep, pigs, and dogs.
Moreover, it can be used for controllable 3D synthesis.
For instance, we can relight our generated 3D assets, swap their textures by conditioning on another input, and animate our articulated shapes, giving us more precise control over the generated assets.
}\label{fig:teaser}
\bigbreak]
\def\thefootnote{*}\footnotetext{Equal contribution.}\def\thefootnote{\arabic{footnote}}
\begin{abstract}
    We present \method, a method for learning category-specific 3D reconstructors for articulated objects, relying solely on ``free'' virtual supervision from a pre-trained 2D diffusion-based image generator.
    Recent approaches can learn a monocular network that predicts the 3D shape, albedo, illumination, and viewpoint of any object occurrence, given a collection of single-view images of an object category.
    However, these approaches heavily rely on manually curated clean training data, which are expensive to obtain.
    We propose a framework that uses an image generator, such as Stable Diffusion, to generate synthetic training data that are sufficiently clean and do not require further manual curation, enabling the learning of such a reconstruction network from scratch.
    Additionally, we incorporate the diffusion model as a score to enhance the learning process.
    The idea involves randomizing certain aspects of the reconstruction, such as viewpoint and illumination, generating virtual views of the reconstructed 3D object, and allowing the 2D network to assess the quality of the resulting image, thus providing feedback to the reconstructor.
    Unlike work based on distillation, which produces a single 3D asset for each textual prompt, our approach yields a monocular reconstruction network capable of outputting a controllable 3D asset from any given image, whether real or generated, in a single forward pass in a matter of seconds.
    Our network can be used for analysis, including monocular reconstruction, or for synthesis, generating articulated assets for real-time applications such as video games.
    The code can be found on the project page at {\small\url{https://farm3d.github.io/}}.
\end{abstract}

\section{Introduction}%
\label{s:intro}

The success of generative AI has led to significant progress in image generation.
Methods like DALL-E~\cite{Ramesh21dalle}, Imagen~\cite{saharia2022imagen}, and Stable Diffusion~\cite{rombach2022stablediffusion} can produce high-quality images from textual prompts.
While these image generators are not trained with 3D capabilities, they can be used to recover 3D shapes by distillation~\cite{poole2022dreamfusion, melas-kyriazi23realfusion}.

A limitation of these distillation methods, is that they require test-time optimization to recover each 3D output.
Not only is this slow, but it also does not learn a true statistical model of the 3D objects.
In this paper, we thus ask whether it is possible to extract from a 2D image generator, not individual 3D instances, but a \emph{feed-forward model for an articulated 3D object category}.
A category-level model, say for cows, sheep or horses, captures an entire family of articulated 3D objects.
Once learned, it can reconstruct a specific object from a single real or generated image in a feed-forward manner, in seconds rather than in hours.

\begin{figure}[t]
    \vspace{-0.10in}
    \centering
    \includegraphics[trim={0 0px 0px 0px}, clip, width=\linewidth]{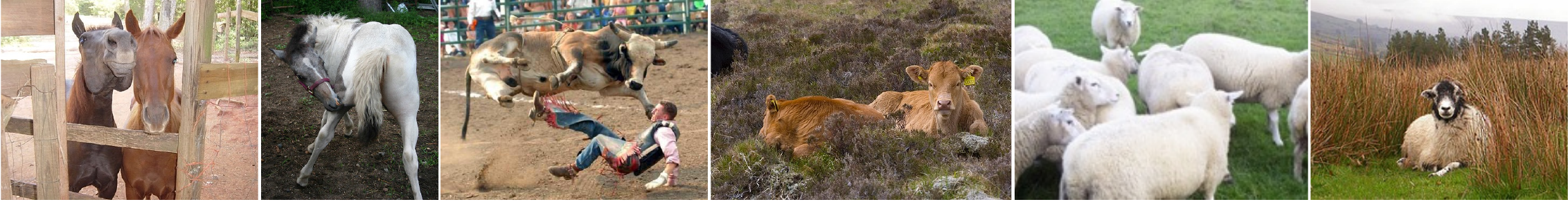}
    \caption{%
    Examples of typical unsuitable images from ImageNet.
    }%
    \label{fig:imagenet}
    \vspace{-0.13in}
\end{figure}

Besides the theoretical interest, answering this question has a practical interest too.
Recent works~\cite{Li2020umr, wu2023magicpony} have learned analogous models of 3D object categories using real images, but they are sensitive to excessive occlusions, truncations, background clutter, and extreme poses.
Such noise is present even in relatively curated datasets like ImageNet (\cite{imagenet}, \cref{fig:imagenet}) -- our estimate is that 41\% of its images contain occlusions -- and manually filtering such images to train one of these models would require about 9 hours per category.\footnote{We assume 2 seconds/image, 41\% of noisy images, aim for $10$k clean images, as in~\cite{wu2023magicpony}.}

In this paper, we show that we can \emph{wholly replace} this real training data with synthetic images obtained from a 2D image generator.
Remarkably, while these generators are in their own right trained on billions of uncurated images, the images that they output are on average much cleaner -- we estimate that only 4\% of Stable Diffusion~\cite{rombach22high-resolution} images that we generate contain occlusions.
For each category, we can thus simply specify a textual prompt\footnote{The prompt is engineered once for all categories.} and obtain clean training data without manual filtering.
Notably, the learned 3D models still generalize to real images at test time.

Besides generating training images, we also extend  Score Distillation Sampling (SDS)~\cite{poole2022dreamfusion} to provide virtual supervision not for a specific 3D object, but for the category-level model itself.
The advantage is that, while the training images only capture a single view of the object, distillation can provide virtual supervision for arbitrary viewpoints, thus providing multi-view constraints that cannot be obtained from monocular real single-view images.
We also make an important technical point and show how to match the energy of the noise in SDS to the learning process in order to improve convergence.

We assess \method qualitatively and quantitatively.
For the latter, we find prior work lacking a dataset with 3D ground truth to be used for direct evaluation of 3D reconstructions, resorting instead to evaluating on 2D proxies like keypoint transfer.
To remedy this problem, we introduce a new synthetic 3D benchmark dataset designed for monocular articulated 3D object reconstructions.
In this way, we are the first to evaluate this class of methods directly on their task of 3D geometry recovery.
Using this benchmark, despite not using any real images for training, thus avoiding tedious data collection and curation, our method achieves performance comparable to state-of-the-art baselines.

To summarise, we make the following contributions:
(1) we show that we can learn models of articulated 3D object categories using only a 2D image generator, without real data;
(2) at test time, these models can reconstruct individual 3D objects from a single real or synthetic image in seconds instead of hours required by distillation;
(3) we show that 2D image generators implicitly ``curate'' billions of noisy training images and produce clean training data without manual curation;
(4) we extend SDS to supervise not individual objects, but a statistical model of an entire category of such objects;
(5) we show how to improve SDS by matching the noise energy to the learning process;
(6) we introduce a new synthetic dataset for a more rigorous evaluation of articulated 3D reconstruction methods.

\section{Related Work}%
\label{s:related}

\paragraph{Weakly-Supervised 3D Object Learning.}

While reconstructing deformable 3D objects typically requires simultaneous multi-view captures~\cite{Hartley2004}, recent works have shown that it is possible to learn 3D models of deformable objects purely from monocular image collections, with a form of geometric supervision in addition to segmentation masks, like keypoint annotations~\cite{Kar15, kanazawa18cmr}, category-specific template shapes~\cite{Goel20ucmr, kulkarni20articulation-aware, kokkinos2021point}, semantic correspondences distilled from image features~\cite{Li2020umr, yao2022lassie, wu2023magicpony, yao2023hi-lassie} and/or strong assumptions like symmetry~\cite{wu20unsupervised, wu2021derender}.
Alternatively, with a known prior viewpoint distribution, a generative adversarial framework can be used to learn simpler 3D objects, such as faces and cars~\cite{nguyen2019hologan, Schwarz2020graf, chan2021piGAN, Chan2022eg3d}.
Monocular videos have also been used for training, incorporating additional temporal signals for learning~\cite{yang21lasr, yang2021viser, wu2021dove, yang2022banmo, yang2023rac}.
Although impressive results have been shown, most still rely on heavily curated category-specific data for training, limiting the model to only few categories.
Here, we introduce a method for distilling 3D objects from large 2D diffusion models, which can potentially generalise to a wide range of object categories.

\begin{figure*}[t!]
\centering
\includegraphics[trim={0px 5px 5px 2px}, clip, width=1\linewidth]{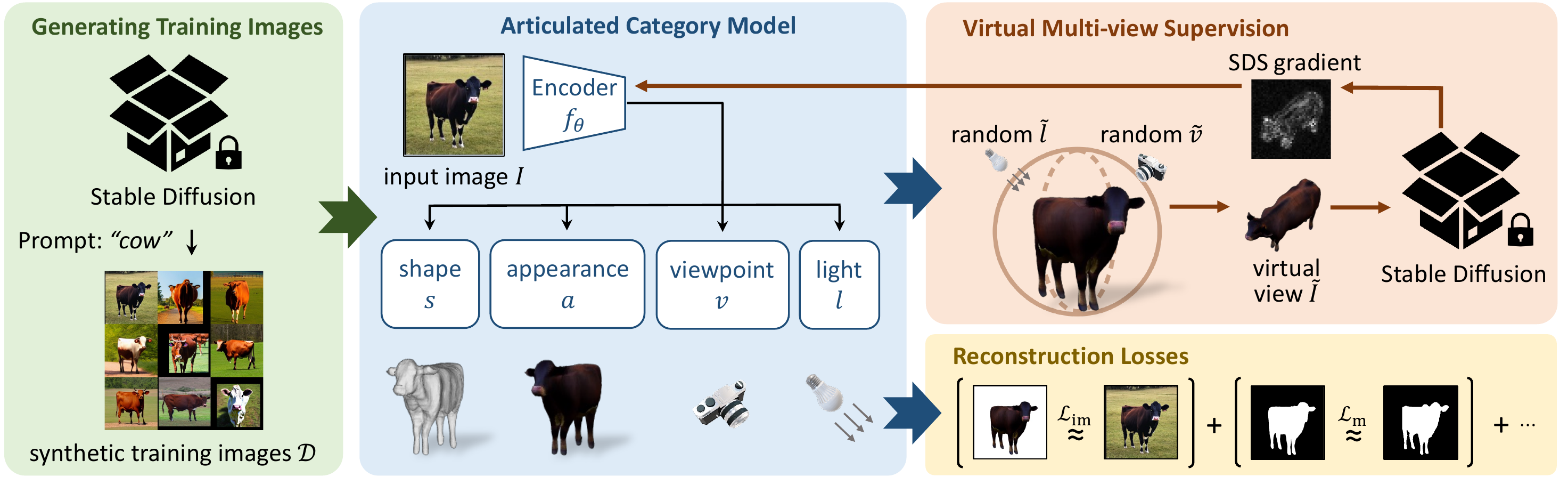}
\caption{\textbf{Training Pipeline.}
We prompt Stable Diffusion for synthetic images of an object category that are then used to train a monocular articulated object reconstruction model that factorises the input image of an object instance into articulated shape, appearance (albedo and diffuse and ambient intensities), viewpoint, and light direction.
During training, we also sample virtual instance views that are then ``critiqued" by Stable Diffusion to guide the learning.
}%
\label{fig:method}
\vspace{-0.12in}
\end{figure*}

\paragraph{Diffusion Models.}

Diffusion models in recent years~\cite{sohl-dickstein15, song2019generative, ho2020denoising, song2021scorebased} became the keystone of text-to-X generative models, where X can be images~\cite{Ramesh21dalle, saharia2022imagen, rombach2022stablediffusion}, videos~\cite{singer2023makeavideo}, vector graphs~\cite{jain2022vectorfusion}, audio~\cite{liu2023audioldm} and so on.
These models generate complex high-fidelity samples by learning to reverse a diffusion process, \ie, 
progressively removing synthetically added noise until the image is recovered.
Text-to-image diffusion models~\cite{Ramesh21dalle, saharia2022imagen, rombach2022stablediffusion}, in particular, introduce textual conditioning to these generative models, which offers a powerful interface for general controllable image generation.
Although these models have demonstrated some level of compositionality and controllability, it is unclear what kind of 3D information is encoded in these learned image synthesis models.
Recently, diffusion models have been investigated for generating training data for 2D tasks~\cite{sariyildiz2023fake,tian2023stablerep}.
Here, we use them to generate training data for learning a category-specific 3D reconstructor.

\paragraph{Distilling 3D Models from Image Diffusion Models.}
Several recent works explored the extraction of 3D information from large pre-trained 2D image diffusion models.
DreamFusion~\cite{poole2022dreamfusion} and Magic3D~\cite{lin2022magic3d} have demonstrated the possibility of generating diverse full 3D objects from text queries by prompting pre-trained image diffusion models.
Make-A-Video3D~\cite{singer2023text4d} adopts a similar strategy for generating 4D dynamic scenes; RealFusion~\cite{melas-kyriazi23realfusion} extends the pipeline to reconstructing 3D objects in photos.

Our method differs from these approaches by learning an articulated category-level model.
This has several advantages.
First, it predicts 3D shapes in a single forward pass, eliminating the need for lengthy optimisation as required by other methods.
Second, the category-level model enables us to directly relate semantically corresponding points on the surface of objects within the same category, which enables numerous applications, such as texture swapping (\cref{fig:synth}) and image understanding (\cref{tab:pascal}).
Third, our method learns articulated shapes, which can be readily animated.

\section{Method}%
\label{s:method}

Our goal is to learn an articulated 3D model of an object category, such as cows, sheep or horses, using exclusively a pre-trained 2D image generator (\cref{s:sd}).
We start from a method designed to learn 3D models from real data (\cref{s:magic-pony}) and
(1) replace the real data with images obtained by applying to a 2D generator to category-agnostic textual prompts (\cref{s:prompting}) and
(2) modify the 3D model's training objective to use the 2D diffusion model as a critic (\cref{s:critic}).
We provide an overview of the method in~\cref{fig:method}.

\subsection{Background: Articulated Category Model}%
\label{s:magic-pony}

We base our model on \magicpony~\cite{wu2023magicpony}, which learns articulated 3D objects from real monocular images.
It does so by training a monocular reconstruction network as a photo-geometric auto-encoder.
This auto-encoder, $f_\theta$, receives as input a single RGB image $I\in\mathbb{R}^{3\times H\times W}$ of the object and outputs a set of photo-geometric parameters $(s, a, v, l) = f_\theta(I)$ describing it.
Here, $\theta$ are the parameters of the autoencoder,
$s$ is the object shape (which incorporates a category-level shape prior, an instance-specific deformation, and the image-specific skeleton pose),
$a$ is the appearance (accounting for albedo and diffuse and ambient illumination),
$v \in SE(3)$ is the object viewpoint (expressed as a rotation and a translation with respect to the camera), and
$l \in \mathbb{S}^2$ is the dominant direction of the illuminant.

The photo-geometric encoder $f_\theta$ is paired with a rendering function
$
\hat I = R(s, a, v, l)
$
which outputs the image of the object.
Key to the method is the fact that $R$ is a \emph{handcrafted} (not learned) differentiable renderer, which implicitly assigns $s, a, v, l$ their photo-geometric meaning.

The model is learned from a collection of single-view images, assuming that images are cropped around the objects with no occlusions or truncation.
Learning also assumes knowledge of the object masks $M \in \{0,1\}^{H\times W}$, obtained using a segmenter such as PointRend~\cite{kirillov2019pointrend}.
Given the resulting set $\mathcal{D}$ of training image-mask pairs $(I,M)$, \magicpony minimises the objective function
$$
\mathcal{L}(\theta|\mathcal{D})
=
\frac{1}{|\mathcal{D}|}
\sum_{(I,M)\in\mathcal{D}}
\mathcal{L}(f_\theta(I) | I, M)
+
\mathcal{R}(f_\theta(I)),
$$
where the term
$
\mathcal{L}(f_\theta(I) | I, M)
$
measures how well the predicted 3D object reconstructs the input image $I$, the input mask $M$, and off-the-shelf image features $\Phi(I)$ (ViT-DINO), and the regulariser
$
\mathcal{R}(f_\theta(I))
$
(Eikonal loss for an SDF) smooths the prior shape and controls the amount of instance-specific deformation and articulation.

\subsection{Background: Image Generation with Diffusion}%
\label{s:sd}

For the image generator we use Stable Diffusion (SD)~\cite{rombach2022stablediffusion}.
SD uses an auto-encoder $h$ to reduce the image $I$ to a smaller latent code $z_0=h(I)\in\mathbb{R}^{D'\times H_h \times W_h}$.
It then learns a conditional distribution $p(z_0|y)$ of the latent code conditioned on a textual prompt $y$.
It does so by considering the sequence of noised signals
$
z_t = \alpha_t z_0 + \sigma_t \epsilon_t = \alpha_t h(I) + \sigma_t \epsilon_t
$
where
$\epsilon_t$ is normally distributed noise,
$\alpha_t = \sqrt{1 - \sigma_t^2}$, and
$\sigma_t \in (0,1)$, $t=0,1,\dots,T$ is an increasing sequence of noise standard deviations (from $\sigma_0\approx0$ to $\sigma_T\approx 1${)}.
The model consists of a `denoising network'
$
\hat \epsilon_t(z_t|y)
$
that approximates the noise content $\epsilon_t$ of $z_t$.
The latter is trained to minimise a loss of the type
\begin{equation}
\mathcal{L}(\hat \epsilon)
=
\mathbb{E}_{t,\epsilon,I,y}
[\| \hat\epsilon_t(\alpha_t h(I) + \sigma_t \epsilon_t|y) - \epsilon_t\|^2 ],
\end{equation}
averaged over noise levels $t$, noise samples $\epsilon$ and empirical samples $(I,y)$ (captioned images).

In order to draw an image sample $I$ from the learned distribution,
one draws a random sample $z_T$ from a normal distribution and then progressively denoise it with
$
z_{t-1} = \alpha_t^{-1}(z_t - \sigma_t \hat\epsilon_t(z_t))
$
to obtain $z_0$ and eventually $I = h^{-1}(z_0)$.

\subsection{Generating Training Images via Prompting}%
\label{s:prompting}

\begin{figure}[t!]
    \centering
    \includegraphics[trim={0 0 0px 0}, clip, width=\linewidth]{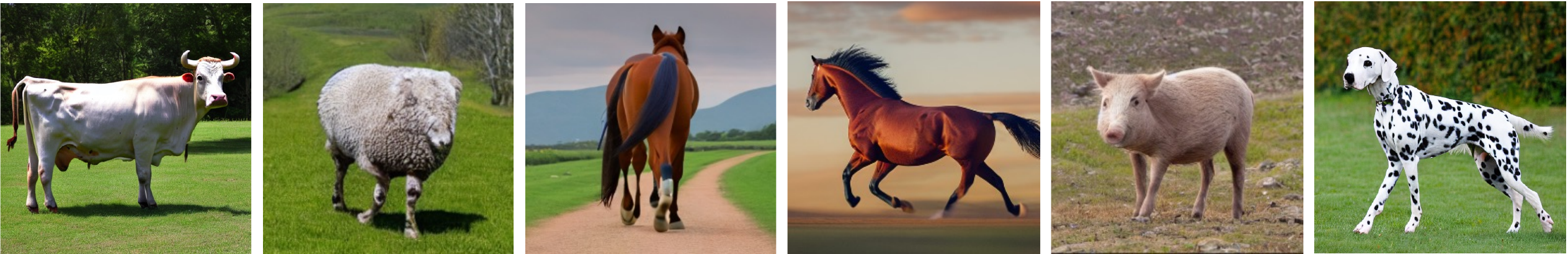}
    \caption{\textbf{Synthetic Training Images Generated with Stable Diffusion.}
    The generated animals are typically without occlusions but sometimes anatomically incorrect (\eg, columns 2 and 3), but our model is robust to this and learns plausible 3D shapes.
    }
    \label{fig:synth-images}
\end{figure}

In~\cref{s:magic-pony}, the training data $\mathcal{D}$ comprises curated real images.
We propose replacing it with \emph{synthetic} images from SD\@, prompted with text specific to each object category, derived from the same category-agnostic textual template, with the only variation being the category name.

When prompted using only the name of the object category, \eg, `cow', SD generates mostly frontal and side views of the object.
We hypothesise that this is due to the bias contained in its training data, an observation similar to that of~\cite{poole2022dreamfusion}.
We found that naive view-dependent prompting~\cite{poole2022dreamfusion} for \{\emph{side, front, back}\} does not work well with SD (as opposed to Imagen~\cite{saharia2022imagen} used in~\cite{poole2022dreamfusion}).
Instead, we found that prompting SD for an animal category ``walking away from the camera'' biases the generation well enough to obtain images that have diverse view coverage.
More details on the textual prompt design together with qualitative analysis are included in the supp.~mat.

\begin{figure}[t!]
\centering
\includegraphics[trim={0px 0px 0 15px}, clip, width=\linewidth]{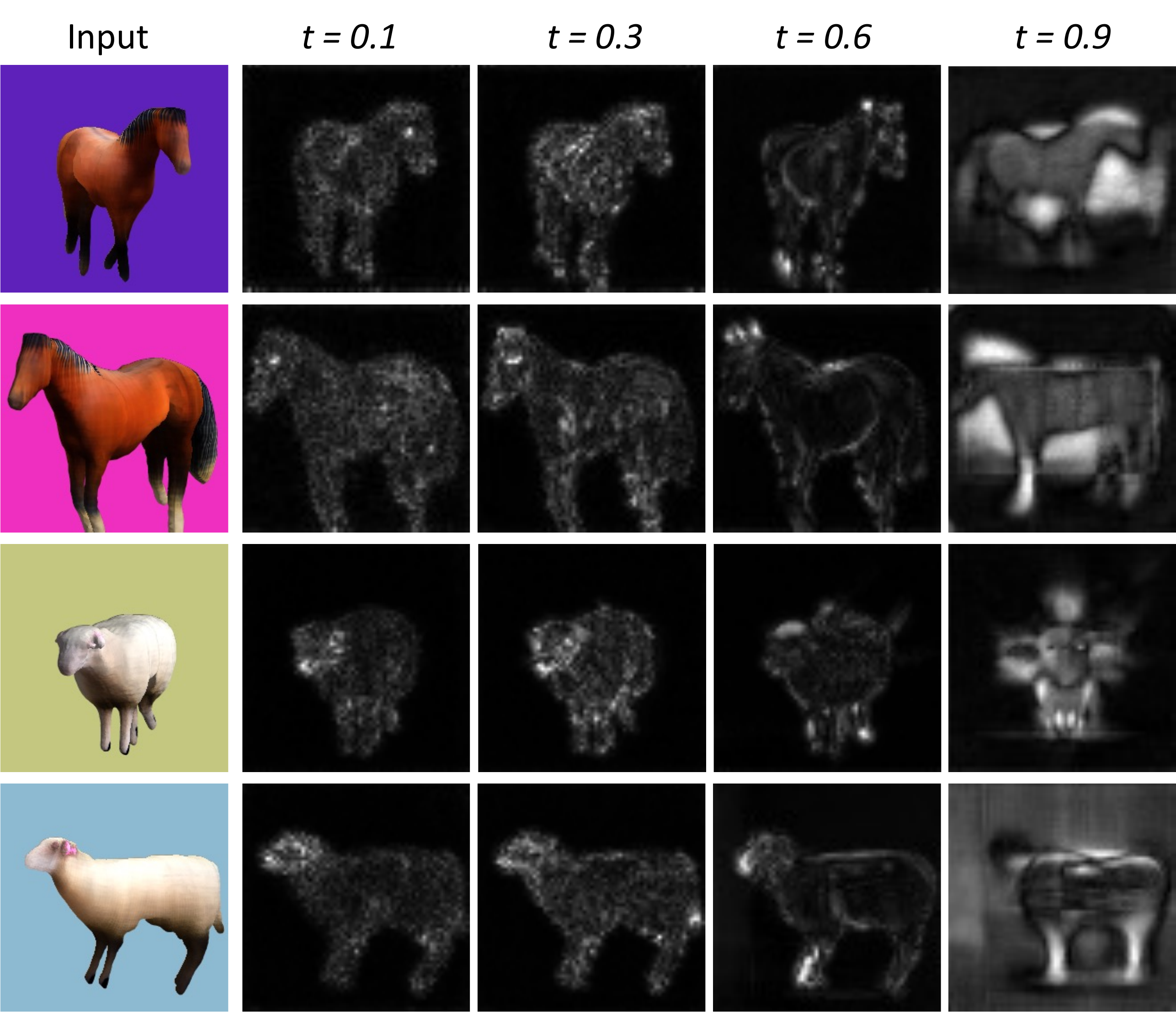}
\caption{\textbf{Noise Scheduling and SDS Gradient.}
We show four rendered images $\hat I$, obtained by first sampling a random viewpoint and illumination as for training our model.
Then, we pick a fixed noise sample $\epsilon$ and show the SDS gradient
$
(\hat \epsilon_t(z_t|y) - \epsilon) ({\partial h}/{\partial \hat I})
$
used to update $\hat I$ in \cref{e:sds} for different values of $\sigma_t$.
Because $\epsilon$ is fixed,
$
z_t = \alpha_t h(\hat I) + \sigma_t \epsilon
$
only depends on $\sigma_t$.
Large noise levels ($t=0.9$) generate an update which is essentially independent of the input image $\hat I$.
Lower noise levels provide more meaningful gradients and lead to more stable training as shown in~\cref{fig:ablation-sds_noise}.
}%
\label{fig:noise}
    \vspace{-0.12in}
\end{figure}

\begin{figure*}[t]
\centering
\includegraphics[trim={10px 8px 25px 0px}, clip, width=1\linewidth]{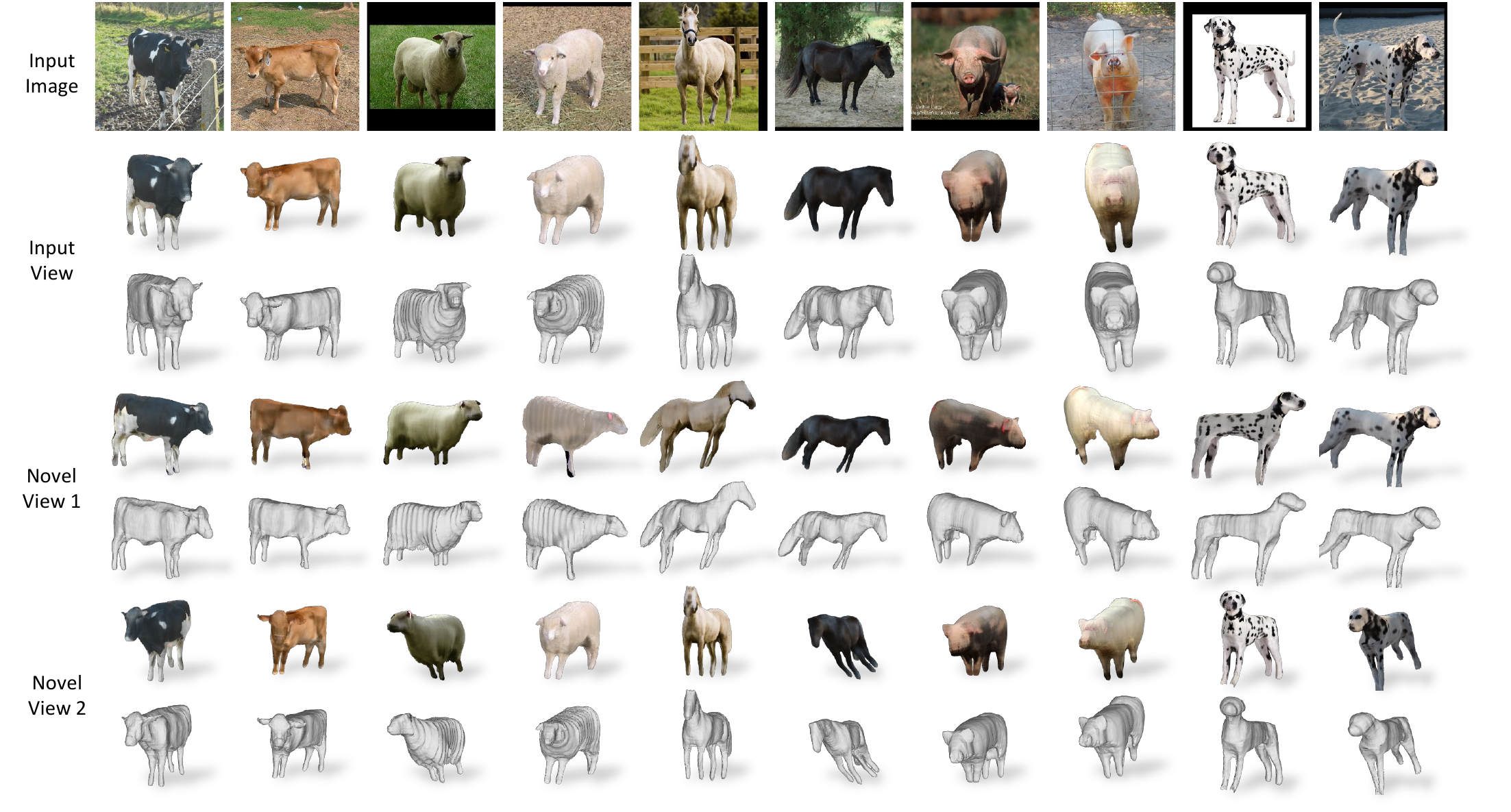}
\caption{\textbf{Single-View Real Image Reconstruction.}
We visualize the predicted textured 3D meshes from both input and novel views.
Our method learns 3D models of a wide range of categories with fine details, \eg legs and ears, despite not being trained on any real images.
}%
\label{fig:recon}
\vspace{-0.12in}
\end{figure*}

\subsection{Distilling a 3D Reconstructor}%
\label{s:critic}

We can use the SD image generator not only to obtain training images but also for \emph{virtual multi-view supervision} to further constrain the reconstruction network $f_\theta$.
Given an example image $I$, we first obtain an estimate  $(s, a, v, l) = f_\theta(I)$ of the object's photo-geometric parameters.
Then, we randomly sample a \emph{new camera viewpoint} $\tilde v$ and a \emph{new light direction} $\tilde l$, and render the corresponding image as follows:
$$
\tilde I
= R(s,a,\tilde v, \tilde l)
= R \circ \operatornamewithlimits{subst}_{\tilde v, \tilde l}
\circ f_\theta(I).
$$
Here the operator
$
\operatorname{subst}_{(\tilde v, \tilde l)}
$
replaces the predicted viewpoint $v$ and light $l$ from the output of $f_\theta(I)$ with the new values $\tilde v$ and $\tilde l$.
We have no direct supervision for the resulting image $\tilde I$, but, we can use the 2D image generator as a critic to judge whether $\tilde I$ `looks correct'.

Concretely, this is achieved through a new way of using the SDS loss~\cite{poole2022dreamfusion}.
Given a training image $I$, the corresponding caption $y$, and randomly-sampled viewpoint $\tilde v$ and lighting $\tilde l$ parameters, the \emph{gradient} of the distillation loss is:
\begin{equation}\label{e:sds}
\nabla_{\theta} \mathcal{L}_{\text{SDS}}
(\theta|\tilde v, \tilde l, I, y) =
\mathbb{E}_{t, \epsilon}
\left[
w_t\cdot
\left(\hat \epsilon_t \left(z_t | y\right)-\epsilon_t\right)
\frac{\partial z_t}{\partial \theta}
\right],
\end{equation}
where $w_t >0$ is a weight factor, $\epsilon_t$ a noise sample, and
$$
z_t = \alpha_t
\cdot
\left(h
\circ R
\circ \operatornamewithlimits{subst}_{\tilde v, \tilde l} 
\circ f_\theta
\right)(I)
+\sigma_t \epsilon_t
$$
is the noised version of the object image obtained from the new viewpoint and lighting.
Based on this definition, the derivative in \cref{e:sds} is, with a slight abuse of notation,
$$
\frac{\partial z_t}{\partial \theta}
=
\alpha_t
\cdot \left.\frac{\partial h}{\partial I}\right|_{\tilde{I}}
\cdot \left.\frac{\partial {R}}{\partial (s,a)}\right|_{(s,a,\tilde v, \tilde l)}
\cdot \left.\frac{\partial [f_{\theta}]_{sa}}{\partial\theta}\right|_I.
$$
The first factor $\alpha_t$ comes from the diffusion scaling.
The second factor is the derivative of the Stable Diffusion latent code $z_0$ with respect to the coded image $I$.
The third factor is the derivative of the rendering function with respect to the shape and appearance components only.
The last factor is the derivative of the shape $s$ and appearance $a$ predictions by the model $f_\theta$.
Note that the viewpoint $\tilde v$ and the lighting $\tilde l$ are not included in the derivative because they are not estimated by $f_\theta$ but sampled.

\paragraph{Noise Scheduling.}

We found that scheduling the noise properly in \cref{e:sds} is critical.
If the noise is too large, the noised latent image bears little to no relation to the one that is actually reconstructed; as a result, the feedback from the SDS loss points in a novel direction, reconstructing `any' object compatible with the prompt $y$ (as in DreamFusion~\cite{poole2022dreamfusion}), instead of improving the reconstruction of the \emph{specific} object contained in the input image $I$.
As motivated in \cref{fig:noise}, in practice, we sample 
$
t \sim \mathcal{U} (0.02, t_\mathrm{max})$ where$ \enspace t_\mathrm{max} = 0.6
$
in the definition of \cref{e:sds}, as opposed to DreamFusion that uses $t_\mathrm{max} = 0.98$.
The choice of $t_\mathrm{max}$ is further supported by an ablation study in \cref{fig:ablation-sds_noise}.

\subsection{Discussion}%
\label{s:discussion}

While our approach borrows several ideas from  DreamFusion~\cite{poole2022dreamfusion}, the application is rather different.
DreamFusion distils from the image generator $\epsilon$ a \emph{single} 3D object $V$ (a neural radiance field), corresponding to a single textual prompt $y$.
We learn instead a \emph{monocular reconstruction network} $f_\theta$ which embodies a category-level articulated 3D model.
At test time, this network can infer a 3D object $(s, a, v, l) = f_\theta(I)$ from a single image $I$ of a new instance in seconds, through feed-forward calculations.
This should be contrasted to inference in DreamFusion, which requires hours of test-time optimization for each 3D object.
Our method can also be used to reconstruct photographs of real objects, which DreamFusion is not capable of, and the output is not a radiance field, but an explicit \emph{articulated mesh}, which can be easily used in applications.

The main shortcoming of our model compared to test-time distillation is the fact that generality is limited to a single category.
Furthermore, we make assumptions about the object categories we learn, including the topology of the articulated skeleton (\eg, 4 legs).

\section{Experiments}
\label{s:experiments}
\begin{figure*}[t]
\centering
\includegraphics[trim={3px 7px 25px 5px}, clip, width=0.9\linewidth]{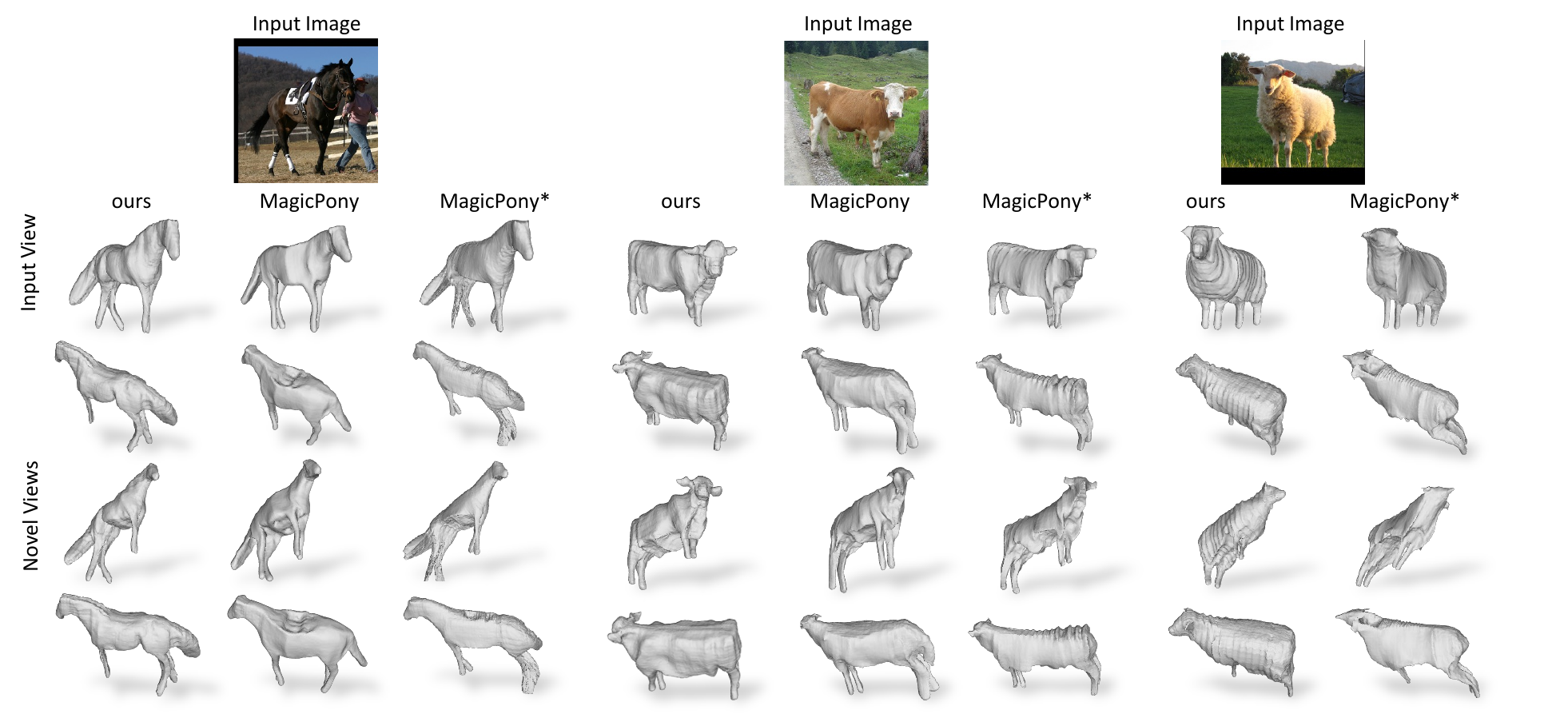}
\caption{\textbf{Qualitative Comparison.}
Comparison with MagicPony~\cite{wu2023magicpony} (Horse and Cow) and MagicPony* trained on ImageNet on real Horse, Cow and Sheep images.
Our model predicts more plausible 3D shapes than MagicPony without being trained on any real images.
When MagicPony* is trained on uncurated data, the predicted shape is less plausible, exhibiting many artefacts and inaccuracies. 
This especially pronounced with cows and sheep, whose training images are noisier than those of horses.
}%
\label{fig:recon-comp}
\vspace{-0.12in}
\end{figure*}

We conduct experiments on both single image 3D reconstruction (\cref{s:reconstruction}) and 3D asset generation (\cref{s:synthesis}) on several animal categories, including cows, sheep, horses, pigs, and Dalmatian dogs, showcasing the generalisation capability of our method across a variety of articulated animal species.
We also compare our approach both qualitatively and quantitatively to prior work, demonstrating that even when training solely on synthetic images, we can still achieve comparable results.

\subsection{Datasets}

We generate $30$k synthetic images per object category using the procedure described in~\cref{s:prompting} to train our method without any real images.
The segmentation masks are obtained with PointRend~\cite{kirillov2019pointrend} and the samples containing likely occluded instances are automatically removed using simple heuristics, as detailed in the sup.~mat.
\Cref{fig:synth-images} shows a few examples of the pre-processed synthetic training data.

For cows, sheep and horses, we evaluate our model qualitatively using PASCAL~\cite{Everingham15} and COCO~\cite{lin2014microsoft} images and quantitatively on the keypoint transfer task using annotations from PASCAL\@.
For pigs and Dalmatian dogs, we show qualitative results using ImageNet~\cite{imagenet} and MagicPony images, respectively. %
Since MagicPony~\cite{wu2023magicpony} does not provide a model for sheep, we collect images from ImageNet sheep synsets to fine-tune their pre-trained horse model.

For all images, 
we follow the same pre-processing procedure as for our synthetic data to obtain segmentation masks and remove truncated objects.

\begin{figure}[!ht]
    \vspace{-0.00in}
    \centering
    \includegraphics[trim={0 0px 0px 0px}, clip, width=\linewidth]{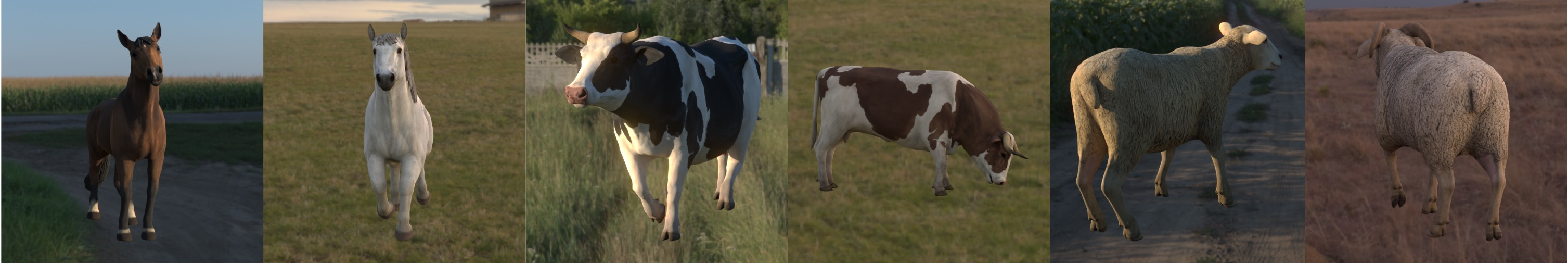}
    \caption{\textbf{
    3D Articulated Animals Dataset.}
    Examples of rendered images from our new dataset used for evaluation.
    }%
    \label{fig:benchmark}
    \vspace{-0.13in}
\end{figure}

\paragraph{Animodel - 3D Articulated Animals Dataset.}
We introduce Animodel, a new 3D articulated animals dataset, to directly evaluate the quality of single-view 3D reconstruction of articulated animals.
This dataset includes textured 3D meshes of articulated animals, including horses, cows, and sheep, crafted by a professional 3D artist, and is accompanied by realistic articulated animations.
For each object category, we generated a randomly sampled set of images, each paired with the corresponding 3D ground-truth mesh.
To render the images, we use Blender\footnote{http://www.blender.org}, and for each asset, we randomly select an animation frame, a viewpoint, and an HDRI environmental map from a publicly available collection\footnote{https://polyhaven.com}, which is used to simulate the background and lighting.
Rendered examples can be found in \Cref{fig:benchmark}.
We intend to release this dataset for future benchmarking purposes.
Additional dataset details are provided in the supplementary material.

\begin{table}[t]
    \setlength{\tabcolsep}{2pt}
    \centering
    \small
    \newcommand{\xpm}[1]{{\tiny$\pm#1$}}
    \caption{\textbf{Quantitative Comparison.}
    We report PCK@0.1 ($\uparrow$ better) on PASCAL VOC and bi-directional Chamfer Distance ($\downarrow$ better), with GT objects normalized in a 1m cube, on 3D Articulated Animals.
    Our method matches MagicPony without training on real images.
    $^\dagger$ A-CSM starts with an articulated 3D model, learning camera and articulations, not shape.
    $^\ddagger$ MagicPony does not offer a model for sheep. Therefore, we fine-tuned their horse model using sheep images from ImageNet.
    * MagicPony trained on ImageNet from scratch.
    Extended version in the sup. mat.
    }
    \resizebox{0.475\textwidth}{!}{
    \begin{tabular}{lrrr@{\hspace{10pt}}rrr}
        \toprule
        \multirow{2}{*}{Method} & \multicolumn{3}{c}{PCK@0.1 (\%) $\uparrow$} & \multicolumn{3}{c}{Chamfer Distance (cm) $\downarrow$} \\
        & Horses & Cows & Sheep & Horses & Cows & Sheep \\
        \midrule
        A-CSM \cite{kulkarni20articulation-aware}$^\dagger$    &  32.9  &  26.3  &  28.6  &  2.93 \xpm{1.16} &  2.35 \xpm{0.68} &  2.48 \xpm{0.70} \\
        \midrule
        UMR \cite{Li2020umr}                                   &  24.4  &  ---    &  --- &  3.51 \xpm{1.51} &  ---    &  ---   \\
        MagicPony \cite{wu2023magicpony}  &  42.9  &  42.5  &  41.2$^\ddagger$  &  2.50 \xpm{0.69} &  2.53 \xpm{0.59} &  3.00 \xpm{0.68}$^\ddagger$ \\
        \midrule
        MagicPony* &  49.1  &  39.9  &  38.3   &  3.48 \xpm{1.06} &  2.54 \xpm{0.62} &  3.93 \xpm{0.62} \\
        Ours (full model)      &  42.5  &  40.2 &  36.1 &  2.85 \xpm{0.83} &  2.41 \xpm{0.54} &  3.31 \xpm{0.49} \\
        Ours w/o SDS  &  41.4  &  37.0 & 37.6  &  3.54 \xpm{1.00} &  2.50 \xpm{0.60} & 3.47 \xpm{0.84} \\
        \bottomrule
    \end{tabular}
    }
    \label{tab:pascal}
\end{table}

\subsection{Single Image 3D Reconstruction}%
\label{s:reconstruction}

\begin{figure}[t]
\centering
\includegraphics[trim={10px 5px 30px 5px}, clip, width=\linewidth]{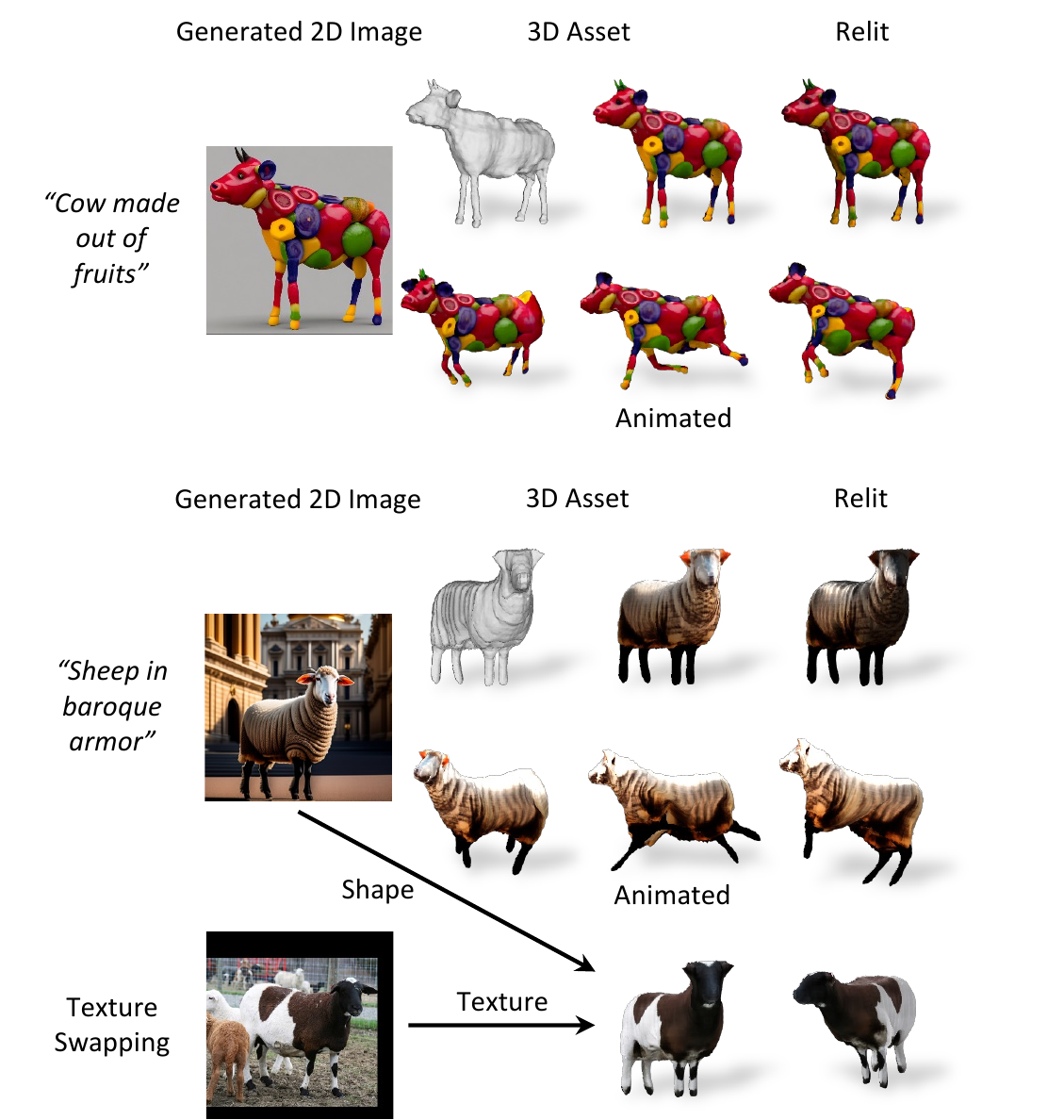}
\caption{\textbf{Controllable 3D Synthesis.}
Our method enables the generation of controllable 3D assets from either a real image or an image synthesised using Stable Diffusion.
Once generated, we have the ability to adjust lighting, swap textures between models of the same category, and even animate the shape.
For shapes out of the training distribution, in this case ``cow made out of fruits" (top row), we use the optional test time shape adaptation (\cref{s:synthesis}).
}%
\label{fig:synth}
\vspace{-0.12in}
\end{figure}

We evaluate the effectiveness of our method on the monocular 3D reconstruction task.
The results, as shown in~\cref{fig:recon}, demonstrate that our method is able to generalise across several animal categories, producing high-quality 3D meshes with faithful texture and accurate articulation, including reconstruction of fine-grained shapes such as legs and ears.
We also provide additional results in the sup.~mat to further support the effectiveness of our approach.

\paragraph{Qualitative Comparisons.}

We compare our model against MagicPony~\cite{wu2023magicpony}, which is trained on a large collection of curated real images, including video frames, whereas our model learns purely from synthetic training data generated by Stable Diffusion.
To further illustrate the vulnerability of weakly-supervised methods to uncurated data, we train MagicPony~\cite{wu2023magicpony} from scratch on ImageNet images from the horse, cow, and sheep synsets, applying the same automatic pre-processing we use for the synthetic images.
\Cref{fig:recon-comp} shows a qualitative comparison of single image 3D reconstruction results.
Despite not seeing any real images at training, our model reconstructs high-quality articulated shapes, as visualized from novel views.
Note that the overall shapes look more natural compared to the MagicPony baseline (\eg no bumps on the horseback), owing to the virtual multi-view supervision from the distillation loss.
This observation is further supported by the ablation study presented in~\cref{fig:ablation-sds}.
The quality of MagicPony's reconstructions significantly deteriorates when trained from scratch using uncurated ImageNet images, which highlights the importance of the time-consuming data curation procedures.

\paragraph{Quantitative Comparisons.}
To evaluate our approach quantitatively, we employ two metrics: keypoint transfer and Chamfer distance between the predicted 3D shape and the ground truth (\cref{tab:pascal}).
Keypoint transfer is commonly used in weakly-supervised object reconstruction~\cite{Li2020umr,kulkarni20articulation-aware,wu2023magicpony}, specifically on PASCAL~\cite{Everingham15}, and it primarily assesses the quality of the predicted pose and articulation.

We also compute Chamfer distance between the predicted and ground-truth shapes using our 3D Articulated Animals Dataset, which provides a direct assessment of the 3D reconstruction accuracy.
We follow the procedure in~\cite{wu2021dove}.
Specifically, we first scale the predicted shape to match the volume of the 3D ground-truth shape and roughly align the canonical pose manually for each method.
We then run the iterative closest point (ICP) algorithm~\cite{besl92a-method} to automatically align each predicted shape to the ground-truth, and report the bi-directional Chamfer distance in cm, with objects normalized into a 1m cube.

In addition to MagicPony~\cite{wu2023magicpony}, we also compare our method with two other weakly-supervised 3D object learning methods: UMR~\cite{Li2020umr} and A-CSM~\cite{kulkarni20articulation-aware}.
Note that only the horse model from UMR has been made publicly available.
UMR does not model complex articulations and, as a result, often produces shapes with reduced accuracy.

Our method significantly outperforms UMR on both metrics.
A-CSM predicts articulated poses given a category-specific articulated template shape, and results in a lower accuracy in the keypoint transfer metric.
On the Chamfer distance metric, it performs reasonably as it relies on an input template shape and learns only the poses.

Even without training on real images, our method achieves comparable results to MagicPony which relies on heavy data curation.
The qualitative comparison, as shown in \cref{fig:recon-comp}, demonstrates a sharp deterioration in MagicPony's performance when trained on uncurated ImageNet.
However, this is not mirrored in the quantitative evaluation.
This is because the keypoint transfer metric only looks at a few points and is not sensitive to artefacts on the surface.
On the Chamfer distance metric, as all existing methods are not yet capable of predicting accurate shapes well aligned with the ground-truth meshes, the errors are still dominated by the inaccurate poses (\eg confusing left and right legs).
Nonetheless, we reach similar performance of prior work by training only on synthetic data.
This also suggests that monocular 3D reconstruction is still far from solved.
We hope the new benchmark will facilitate assessing the progress in this field.

\subsection{Controllable 3D Shape Synthesis}%
\label{s:synthesis}

In addition to controlling articulation via the underlying skeleton model, our approach enables controllable 3D synthesis, as demonstrated in \cref{fig:synth}.
The model can be conditioned on either an image or text.
For text conditioning, such as ``sheep in baroque armor'', Stable Diffusion is first used to generate an example image of the input prompt.
The image is then fed into the model as done with real images.

Unlike distillation approaches that require hours of test-time optimisation, our approach produces the 3D shape in a single pass, enabling faster iterations when producing 3D assets.
Notably, we can explicitly control articulation, lighting, and albedo (\ie, texture swapping) using our method.

\paragraph{Test Time Shape Adaptation (Optional).}

During test time, we can optionally fine-tune our category model on a single image.
This allows us to adapt the model to shapes that are out of the training distribution, such as ``cow made out of fruits'' as demonstrated in \cref{fig:synth}.
As this process is very fast (less than 30 seconds) this allows users to control the outputs of generative models in 3D, which is an exciting capability with many potential applications.

\subsection{Ablations}%
\label{s:ablations}

\begin{figure}[!ht]
    \centering
    \includegraphics[trim={0 8px 27px 6px}, clip, width=\linewidth]{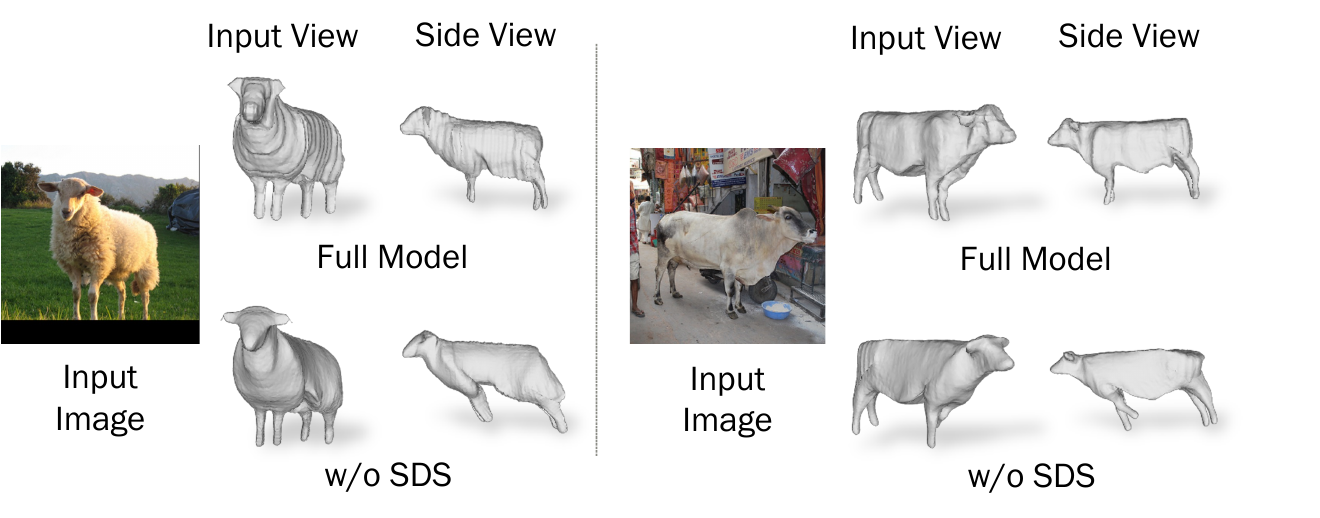}
    \caption{\textbf{Ablation Study on the SDS Loss.}
    The SDS loss utilizes rendered virtual views of an object instance under random directional light.
    These views are encouraged to resemble real images, which helps to achieve more consistent object pose and articulation.
    However, this  may lead to side effects, such as simulating a sheep's 'fluffiness' through surface wrinkles.
    }%
    \label{fig:ablation-sds}
    \vspace{-0.12in}
\end{figure}

We conduct ablations to evaluate the effect of the SDS guidance on the quality of the learned pose and shape, both  quantitatively in~\cref{tab:pascal} and qualitatively in~\cref{fig:ablation-sds}.
Our qualitative results illustrate that using SDS helps the method learn more plausible shapes and articulation.

\begin{figure}[!ht]
    \centering
    \includegraphics[trim={10 7px 25px 5px}, clip, width=\linewidth]{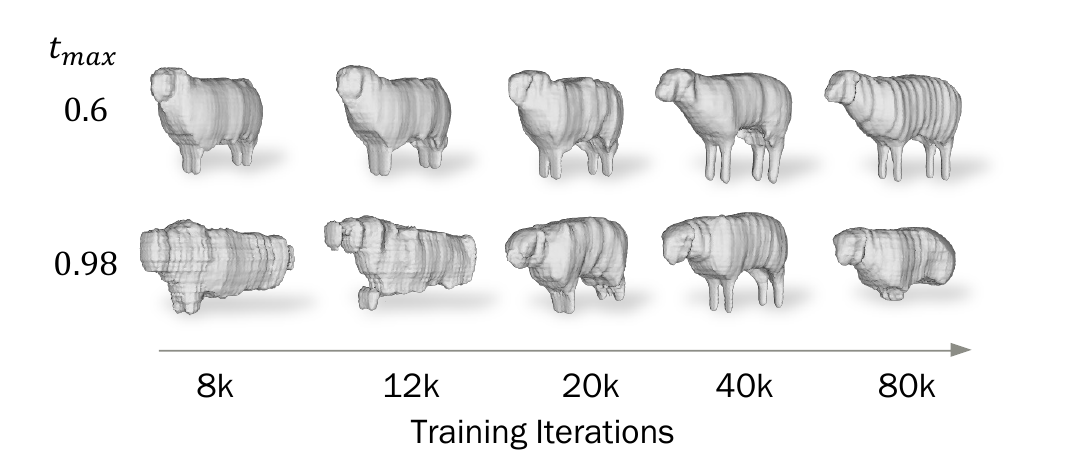}
    \caption{\textbf{Effect of our SDS Noise Scheduling on the Category-Specific Prior Shape during Training.}
    Sampling with $t_{max} = 0.6$ helps the SDS guidance stay more faithful to the input image, and stabilises training as can be seen from the learned prior shape.}%
    \label{fig:ablation-sds_noise}
    \vspace{-0.12in}
\end{figure}

We also perform an ablation study on the noise scheduling in~\cref{fig:ablation-sds_noise}, demonstrating that it is critical to training stability. Without our noise scheduling, the learning is unstable and often collapses multiple times during training.

\section{Conclusions}%
\label{s:conclusions}

We introduced \method, a method for learning a monocular 3D reconstructor of an object category entirely from synthetic data from an off-the-shelf text-to-image generator.
There are many enticing aspects of this approach:
it reduces the need for collecting and curating real data for learning 3D objects, it learns 3D reconstruction networks that produce outputs in seconds at test time, and these networks can be used for monocular reconstruction (analysis) or for 3D asset generation (synthesis), producing articulated meshes ready for animation and rendering in applications.

\paragraph{Data Ethics.}

We use the DOVE, Weizmann Horse, PASCAL, Horse-10, and MS-COCO datasets in a manner compatible with their terms.
Some of these images may accidentally contain faces or other personal information, but we do not make use of these images or image regions.
For further details on ethics, data protection, and copyright please see {\small\url{https://www.robots.ox.ac.uk/~vedaldi/research/union/ethics.html}}.

\paragraph{Acknowledgements.}
Shangzhe Wu is supported by a research gift from Meta Research.
Andrea Vedaldi and Christian Rupprecht are supported by ERC-CoG UNION 101001212.
Tomas Jakab and Christian Rupprecht are (also) supported by VisualAI EP/T028572/1.

{
    \small
    \bibliographystyle{ieeenat_fullname}
    \bibliography{vedaldi_specific,ref,vedaldi_general}
}
\newpage
\appendix
\section*{Appendix}

\begin{table*}[!htb]
    \setlength{\tabcolsep}{2.5pt}
    \centering
    \small
    \newcommand{\xpm}[1]{{\tiny$\pm#1$}}
    \caption{\textbf{Evaluation on 3D Articulated Animals Dataset.}
    We report bi-directional Chamfer Distance in cm ($\downarrow$ better) on the 3D Articulated Animals dataset.
    We evaluate the methods using both the normalized versions of the objects (GT objects fitted inside a 1m cube) and their real-sized counterparts.
    Our method achieves results comparable to those of MagicPony without training on any real images.
    Test-time optimization-based methods directly optimize the test images; in contrast, our method is not trained on these images and reconstructs the 3D shape in seconds.
    $^\dagger$ A-CSM starts from an existing articulated 3D model of the object category and only learns camera and articulations, but not shape.
    $^\ddagger$ MagicPony does not offer a model for sheep. Therefore, we fine-tuned their horse model using sheep images from ImageNet.
    * MagicPony trained on ImageNet from scratch.
    }
    \resizebox{0.999\textwidth}{!}{
    \begin{tabular}{lrrr@{\hspace{10pt}}rrr@{\hspace{10pt}}rrr@{\hspace{10pt}}rrr@{\hspace{10pt}}rrr}
        \toprule
        \multirow{3}{*}{Method} & \multicolumn{6}{c}{1k samples (full dataset)} & \multicolumn{6}{c}{30 samples (subset)} \\
         & \multicolumn{3}{c}{Normalized} & \multicolumn{3}{c}{Real-Sized} & \multicolumn{3}{c}{Normalized} & \multicolumn{3}{c}{Real-Sized} \\
        & Horses & Cows & Sheep & Horses & Cows & Sheep & Horses & Cows & Sheep & Horses & Cows & Sheep \\
        \midrule
        \multicolumn{13}{c}{\emph{Test-Time Optimization-Based Methods}} \\
        \addlinespace[0.2em]
        LASSIE~\cite{yao2022lassie}    & --- & --- & ---  & --- & --- & --- & 2.66 \xpm{0.80} & 2.17 \xpm{0.76} & 2.34 \xpm{0.63} & 7.45 \xpm{2.23} & 6.15 \xpm{1.82} & 2.68 \xpm{0.65} \\
        RealFusion~\cite{melas-kyriazi23realfusion}   & ---  & --- & --- & --- & --- & --- & 5.07 \xpm{1.35} & 4.28 \xpm{0.79} & 5.62 \xpm{1.47} & 14.16 \xpm{3.62} & 12.29 \xpm{2.17} & 6.43 \xpm{1.50} \\
        \addlinespace[0.5em]
        \midrule
        \multicolumn{13}{c}{\emph{Learning-Based Methods}} \\
        \addlinespace[0.2em]
        A-CSM \cite{kulkarni20articulation-aware}$^\dagger$    &  2.93 \xpm{1.16}  &  2.35 \xpm{0.68}  &  2.48 \xpm{0.70} & 8.15 \xpm{3.09} & 6.71 \xpm{1.81} & 2.84 \xpm{0.77} & 2.77 \xpm{1.11} & 2.52 \xpm{0.90} & 2.49 \xpm{0.66} & 7.74 \xpm{2.97} & 7.14 \xpm{2.10} & 2.85 \xpm{0.70} \\
        \midrule
        UMR \cite{Li2020umr}                                  &  3.51 \xpm{1.51}  &  ---    &  ---  & 9.75 \xpm{4.11} & --- & --- & 3.23 \xpm{1.09} & --- & --- & 9.05 \xpm{3.02} & --- & --- \\
        MagicPony \cite{wu2023magicpony}  &  2.50 \xpm{0.69}  &  2.53 \xpm{0.59}  &  3.00 \xpm{0.68} & 6.95 \xpm{1.90} & 7.22 \xpm{1.53} & 3.43 \xpm{0.73} & 2.39 \xpm{0.52} & 2.56 \xpm{0.64} & 2.87 \xpm{0.48} & 6.72 \xpm{1.62} & 7.30 \xpm{1.45} & 3.29 \xpm{0.46} \\
        \midrule
        MagicPony*  &  3.48 \xpm{1.06}  &  2.54 \xpm{0.62}  &  3.93 \xpm{0.62} & 9.69 \xpm{2.85} & 7.23 \xpm{1.56} & 4.49 \xpm{0.62} & 3.34 \xpm{1.13} & 2.63 \xpm{0.79} & 3.77 \xpm{0.66} & 9.35 \xpm{3.15} & 7.46 \xpm{1.74} & 4.32 \xpm{0.65} \\
        Ours (full model)     &  2.85 \xpm{0.83}  &  2.41 \xpm{0.54}  &  3.31 \xpm{0.49} & 7.92 \xpm{2.13} & 6.91 \xpm{1.49} & 3.79 \xpm{0.55} & 2.63 \xpm{0.51} & 2.57 \xpm{0.63} & 3.32 \xpm{0.48} & 7.37 \xpm{1.44} & 7.34 \xpm{1.56} & 3.80 \xpm{0.46} \\
        Ours w/o SDS   &  3.54 \xpm{1.00}  &  2.50 \xpm{0.60} & 3.47 \xpm{0.84} & 9.85 \xpm{2.56} & 7.12 \xpm{1.51} & 3.96 \xpm{0.90} & 3.44 \xpm{1.04} & 2.59 \xpm{0.70} & 3.30 \xpm{0.72} & 9.64 \xpm{2.84} & 7.36 \xpm{1.57} & 3.78 \xpm{0.74} \\
        \bottomrule
    \end{tabular}
    }
    \label{tab:chamfer_big}
\end{table*}

\begin{table}[htbp]
    \centering
    \small
    \caption{\textbf{Evaluation on Keypoint Transfer.}
    We report PCK@0.1 ($\uparrow$ better) on PASCAL VOC.
    Our method achieves results comparable to those of MagicPony without training on any real images.
    RealFusion does not output 3D meshes in canonical space and thus cannot be used for keypoint transfer.
    $^\dagger$ A-CSM starts from an existing articulated 3D model of the object category and only learns camera and articulations, but not shape.
    $^\ddagger$ MagicPony does not offer a model for sheep.
    Therefore, we fine-tuned their horse model using sheep images from ImageNet.
    * MagicPony trained on ImageNet from scratch.
    }
    \begin{tabular}{lrrr}
        \toprule
        Method & Horses & Cows & Sheep \\
        \midrule
        \multicolumn{4}{c}{\emph{Test-Time Optimization-Based Methods}} \\
        \addlinespace[0.2em]
        LASSIE~\cite{yao2022lassie} & 42.2 & 37.5 & 27.5 \\
        \midrule
        \multicolumn{4}{c}{\emph{Learning-Based Methods}} \\
        \addlinespace[0.2em]
        A-CSM \cite{kulkarni20articulation-aware}$^\dagger$ & 32.9 & 26.3 & 28.6 \\
        \midrule
        UMR \cite{Li2020umr} & 24.4 & --- & --- \\
        MagicPony \cite{wu2023magicpony} & 42.9 & 42.5 & 41.2$^\ddagger$ \\
        \midrule
        MagicPony* & 49.1 & 39.9 & 38.3 \\
        Ours (full model) & 42.5 & 40.2 & 36.1 \\
        Ours w/o SDS & 41.4 & 37.0 & 37.6 \\
        \bottomrule
    \end{tabular}
    \label{tab:pascal_lassie}
\end{table}

\begin{figure}[bp]
    \centering
    \includegraphics[trim={0px 0px 30px 0px}, clip, width=\linewidth]{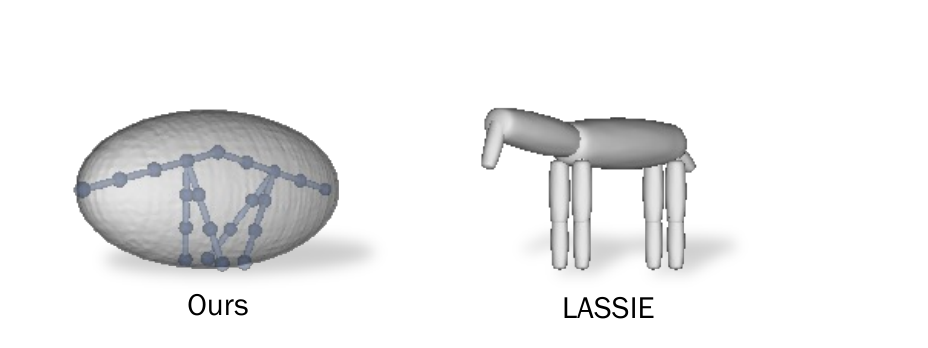}
    \caption{\textbf{Comparison of shape initialisation against LASSIE.}
    Our model begins with a generic ellipsoid with a simple heuristic description of the bone topology, whereas LASSIE~\cite{yao2022lassie} initiates with a skeleton featuring bones initialised as cylinders which are reminiscent of the final object structure.
    LASSIE employs a much stronger hand-crafted initial shape, which simplifies their optimisation process.
    }%
    \label{fig:prior}
\end{figure}

\begin{figure*}[t]
    \centering
    \includegraphics[trim={0px 5px 30px 0px}, clip, width=\linewidth]{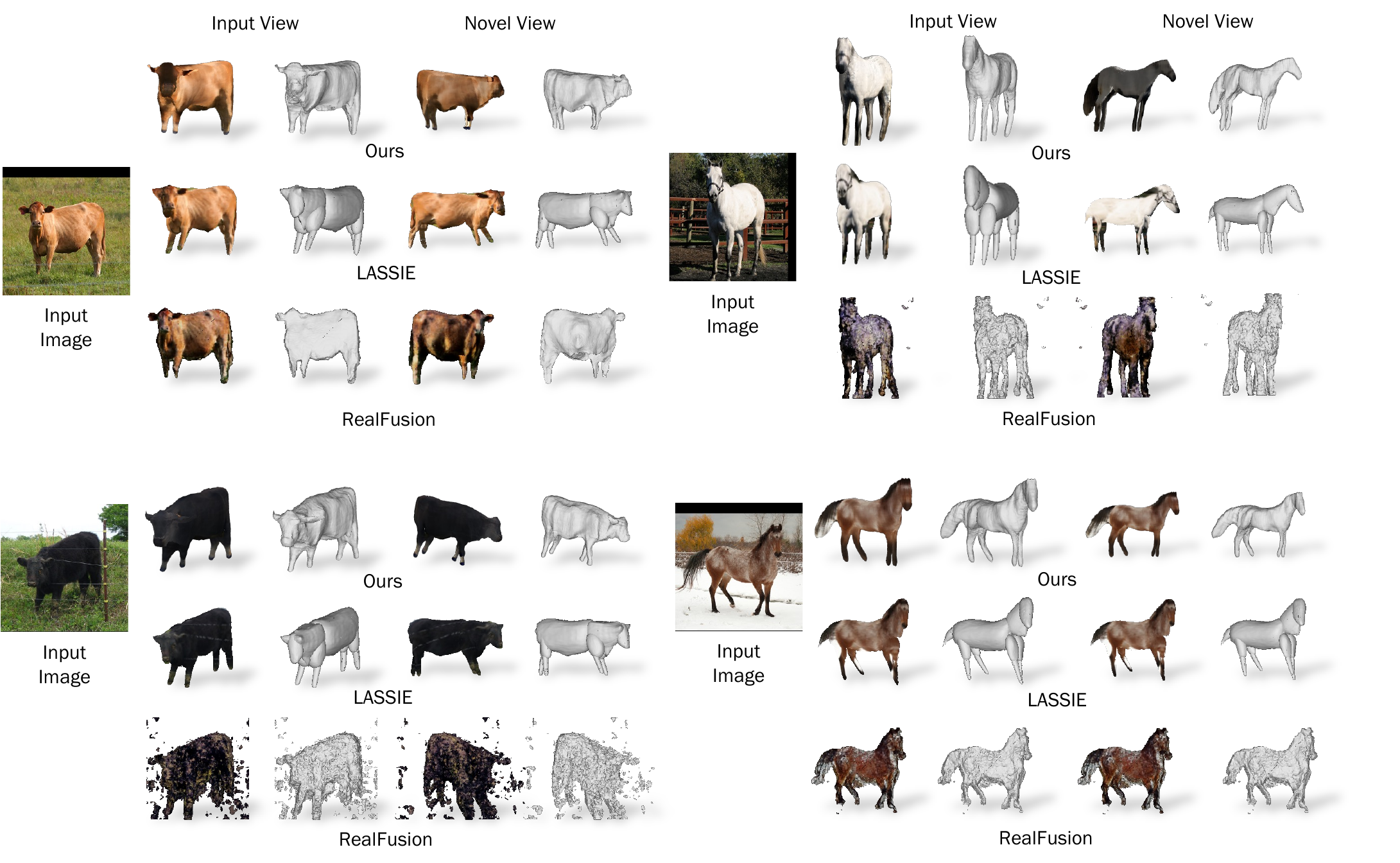}
    \caption{\textbf{Comparison with LASSIE and RealFusion.}
    Our model outputs more realistic 3D shapes than LASSIE~\cite{yao2022lassie} and RealFusion~\cite{melas-kyriazi23realfusion} without \textit{optimising} the shape on the exact same input images.
    Note that when compared to LASSIE our model disentangles lighting from albedo when predicting appearance, resulting in higher-quality texture when the shape is rendered from a novel view -- in the top row, the texture is darker from the un-illuminated side.
     RealFusion tends to predict incorrect geometry that can be flat, full of holes, or with incorrect number of limbs similarly to DreamFusion.
    }%
    \label{fig:lassie-comp}
\end{figure*}

This supplementary material contains extended comparisons with prior work (\cref{s:compare_mp,s:compare_test_time,s:compare_dreamfusion}), a failure case analysis (\cref{s:fail}), and more technical details (\cref{s:details}), which also include additional results with automatically obtained segmentation masks (\cref{s:automask}).

\section{Additional Results}\label{s:additional}

\subsection{Extended Comparison}\label{s:compare_mp}
\begin{figure*}[t]
\centering
\includegraphics[trim={3px 5px 20px 5px}, clip, width=1\linewidth]{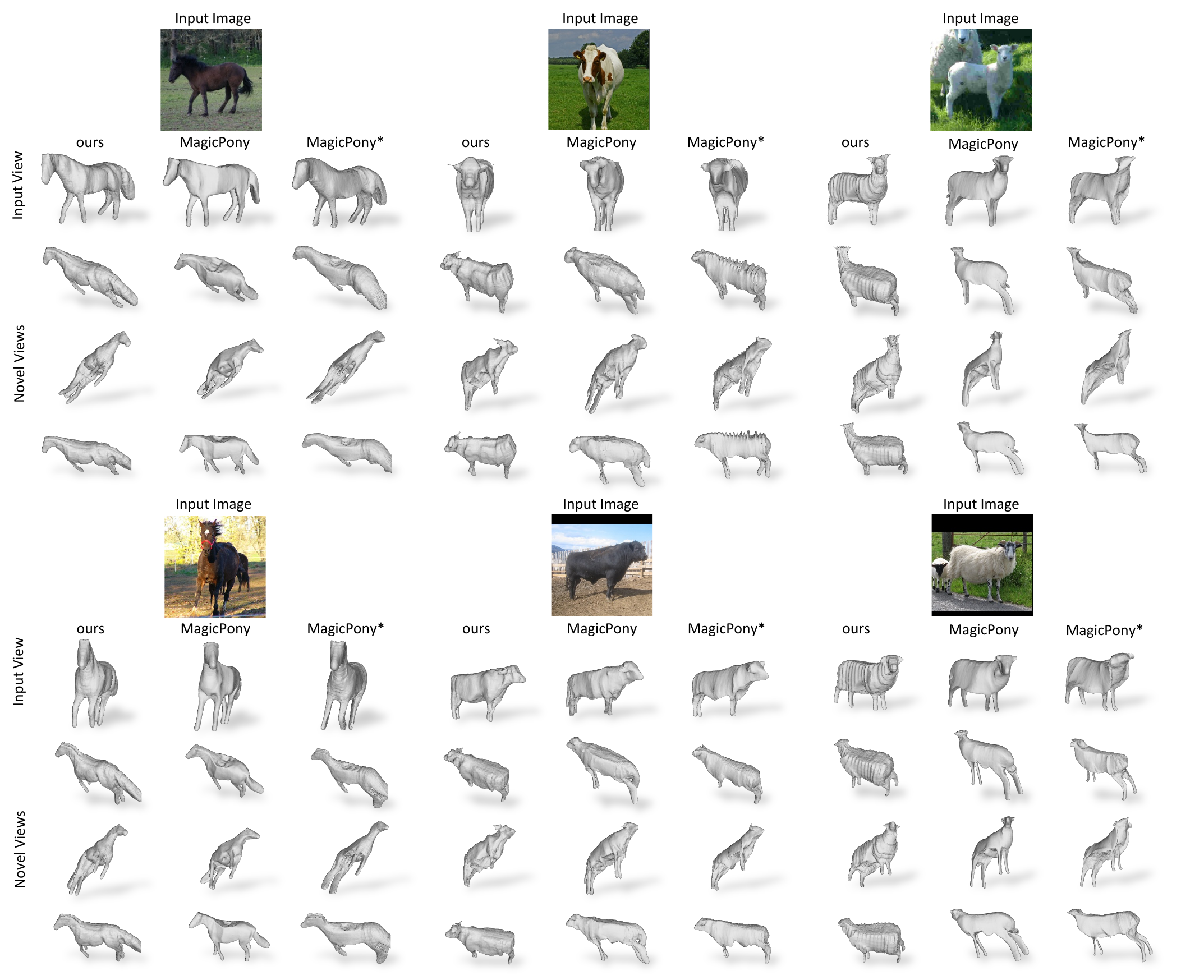}
\caption{\textbf{Qualitative Comparison.}
Comparison with MagicPony~\cite{wu2023magicpony} (Horse and Cow) and MagicPony* trained on ImageNet on real Horse, Cow and Sheep images.
Our model predicts more plausible 3D shapes than MagicPony without being trained on any real images.
When MagicPony* is trained on uncurated data, the predicted shape is less plausible, exhibiting many artefacts and inaccuracies. 
This especially pronounced with Cows and Sheep, whose training images are noisier than those of Horses.
}%
\label{fig:supmat-recon-comp}
\vspace{-0.12in}
\end{figure*}

We include an extended version of quantitative comparison on 3D Articulated Animals Dataset in~\cref{tab:chamfer_big}.
Additionally, we provide extended qualitative comparisons with prior work of MagicPony~\cite{wu2023magicpony} and MagicPony trained on uncurated images from ImageNet in~\cref{fig:supmat-recon-comp}, showcasing the vulnerability of weakly-supervised methods to uncurated data.

\subsection{Comparison with Test-time Optimization Methods}\label{s:compare_test_time}
We also compare our method with recent test-time optimization-based methods, namely RealFusion~\cite{melas-kyriazi23realfusion} and LASSIE~\cite{yao2022lassie}.
A later iteration of LASSIE, Hi-LASSIE~\cite{yao2023hi-lassie}, does not have publicly available code, which prevents us from evaluating against it.
As LASSIE optimizes over an image ensemble consisting of a few (10-30) images, we use a subset of 30 images for each category from the 3D Articulated Animals dataset.
Since RealFusion optimizes over a single image and thus is computationally expensive, we also use the same subset of 30 images.

We show qualitative comparisons in~\cref{fig:lassie-comp}, and report quantitative comparisons in~\cref{tab:pascal_lassie} and~\cref{tab:chamfer_big}.
The key advantage of our method is that it is orders of magnitude faster than test-time optimization-based methods, producing a 3D shape in seconds, whereas LASSIE takes roughly an hour and RealFusion takes roughly $1.5\text{-}3$ hours for each image.

\paragraph{RealFusion.}
RealFusion~\cite{melas-kyriazi23realfusion} extends DreamFusion~\cite{poole2022dreamfusion} by conditioning on an input image, but it inherits similar shortcomings, namely, a long optimization time and not learning the 3D mesh in canonical space, which is crucial for downstream tasks such as keypoint transfer. Additionally, it requires an initial textual inversion step that adds another $60\text{-}120$ minutes to the optimization time, bringing the total time needed to reconstruct shape to $1.5\text{-}3$ hours. Both the qualitative and quantitative results demonstrate that the method is highly unstable and produces shapes with highly distorted geometry.

\paragraph{LASSIE.}
LASSIE~\cite{yao2022lassie} deforms geometric primitive shapes (spheres, cylinders, cones, etc.) for each body bone/part, yielding unnatural bone junctions, as demonstrated in \cref{fig:lassie-comp}.
The initialization also requires more human input than our method, as detailed in \cref{fig:prior}.
Note that LASSIE directly optimizes the reconstruction for each input image, whereas our model has not seen any of these input images during training (nor any real images), showcasing its superb generalizability.

This also highlights the benefit of learning-based methods in unsupervised single-view 3D reconstruction.
Given that this problem is highly ill-posed and under-constrained, optimization-based approaches necessitate strong regularization, which can limit their ability to generate complex shapes.
Conversely, learning-based methods can capitalize on multi-view knowledge acquired from their training data.
This work demonstrates that one can gain comparable, if not better, multi-view prior knowledge from pre-trained 2D diffusion models, which reduces the need to collect datasets for 3D training and boosts training stability.

\subsection{Comparison with DreamFusion}\label{s:compare_dreamfusion}
\begin{figure}[h]
    \centering
    \includegraphics[trim={10px 10px 30px 00px}, clip, width=\linewidth]{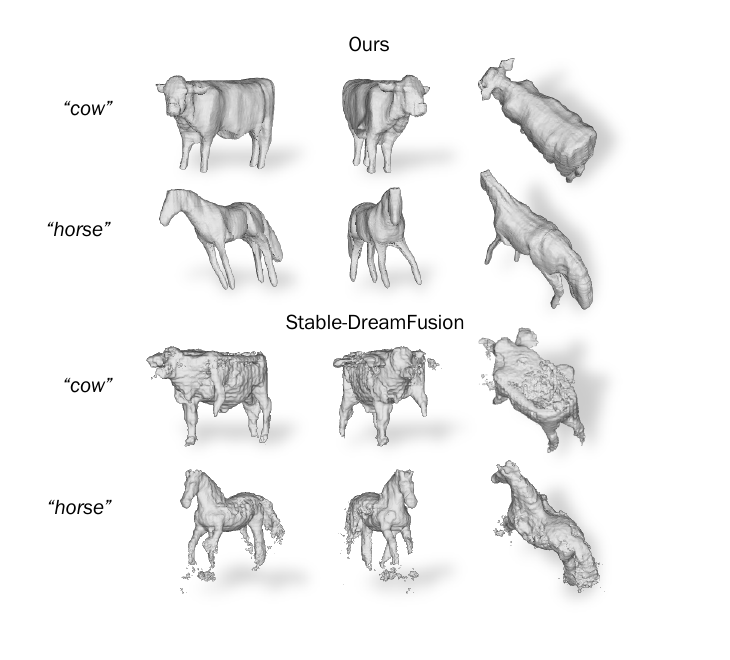}
    \caption{\textbf{Comparison with DreamFusion.}
    We run Stable-DreamFusion~\cite{stable-dreamfusion}, an open-source implementation of DreamFusion~\cite{poole2022dreamfusion}.
    Stable-DreamFusion produces shapes with holes and wrong anatomies (for `cow', half of the left front leg is missing; for `horse', there are 5 legs).
    }%
    \label{fig:dreamfusion-comp}
\end{figure}

To elaborate on how our method can be a preferable candidate for 3D asset generation over category-agnostic text-to-3D methods such as DreamFusion~\cite{poole2022dreamfusion}, we provide an additional qualitative comparison with Stable-DreamFusion~\cite{stable-dreamfusion}, an open-source implementation of DreamFusion that uses Stable Diffusion~\cite{rombach2022stablediffusion} (rather than the non-public Imagen~\cite{saharia2022imagen}) as the 2D prior.
Given a prompt, for Stable-DreamFusion, we extract the mesh from the \textit{optimized} radiance field, using its implementation out of the box; for our method, we first sample 4 images from Stable Diffusion conditioned on the same prompt and remove the images which are likely to contain truncated instances (using the same automatic pipeline as for pre-processing our virtual training images), and then randomly select one image to feed into our category-specific model to get the mesh. We show the generated 3D contents, with their corresponding prompts, in \cref{fig:dreamfusion-comp}. While Stable-DreamFusion takes $30\text{-}60$ minutes to produce the mesh, our method uses less than $10$ seconds ($5$ seconds for generating and pre-processing images and $1$ second for a single forward pass to predict the shape).
In addition, users have more control over the synthesized 3D asset when using our method, as our model can be conditioned on images of the same category; by contrast, even to get a different mesh with (Stable-)DreamFusion, one has to tweak the prompt, change the random seed, and/or modify the guidance scale and wait for another $30\text{-}60$ minutes.

\subsection{Failure Cases}\label{s:fail}
\begin{figure}[t]
    \centering
    \includegraphics[trim={0px 0px 20px 0px}, clip, width=\linewidth]{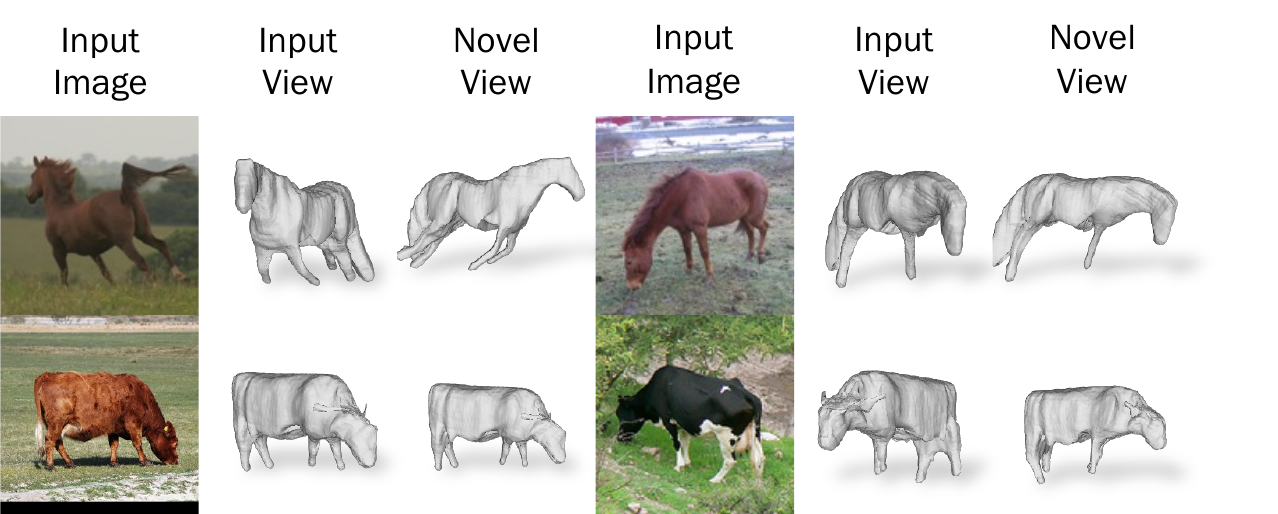}
    \caption{\textbf{Failure Cases.}
    Typical failure modes include incorrect viewpoint prediction and inaccurate articulation.
    }%
    \label{fig:fail}
\end{figure}

\begin{figure}[t]
    \centering
    \includegraphics[trim={0px 5px 0px 0px}, clip, width=\linewidth]{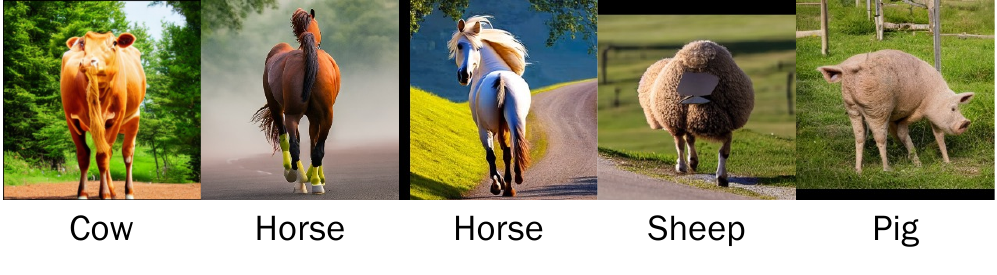}
    \caption{\textbf{Stable Diffusion Failure Cases.}
    Stable Diffusion can generate anatomically wrong images.
    We observe that this is typically correlated with which is particularly significant for instances facing away.
    }%
    \label{fig:wrong}
\end{figure}

As shown in \cref{fig:fail}, our method is more likely to predict wrong viewpoints when the animal instance in the input image is facing away from the camera.
We hypothesize that this is due to two key reasons.
First, Stable Diffusion sometimes generates anatomically incorrect images (examples are shown in \cref{fig:wrong}), especially when the instance is facing away.
Second, even with our engineered prompts, instances facing away from the camera are still underrepresented in the \textit{virtual} training set.
In the case of extreme articulation, the viewpoint and bone predictions are also less stable similarly to MagicPony~\cite{wu2023magicpony}.

\section{Additional Technical Details}\label{s:details}

\subsection{Virtual Multi-View Supervision}
Computing the SDS gradient, Eq. 2 in the main paper, has a relatively large memory footprint and requires a halved batch size, which deteriorates training stability.
We found it helpful in practice to apply SDS guidance every second iteration.
Specifically, at every second iteration, camera positions $\tilde{v}$ are randomly sampled in spherical coordinates, with an elevation angle $\phi_{\text{cam}} \in \left[-10^{\circ}, 90^{\circ}\right]$, an azimuth angle $\theta_{\text{cam}} \in \left[0^{\circ}, 360^{\circ}\right]$, and a distance from the origin in $\left[9, 11\right]$. For reference, we initialize the prior shape as an ellipsoid with axis lengths 2.1, 2.1 and 1.05. We found that randomly sampling the camera field of view (as in DreamFusion) did not improve results. Instead of sampling point lights (as in DreamFusion), we assume a distant directional light following \magicpony, and sample a random light direction $\tilde{l} \sim \mathcal{N}({\tilde{v}}/{\norm{\tilde{v}}}, \textbf{I})$, where $\tilde{v} \in \mathbb{R}^3$ is the sampled camera viewpoint. This sampling strategy, first used in DreamFusion, ensures that the side of the object facing towards the camera is mostly illuminated.

We render $256 \times 256$ images
and, like in DreamFusion, randomly alternate between a shaded texture, texture-less shading, and albedo-only texture (\ie, no lighting).
We also use the same classifier-free guidance (with a classifier guidance strength of 100) as it improves the regularity of the learned shape.

\subsection{Optional Texture Finetuning}
We follow the implementation of MagicPony~\cite{wu2023magicpony} for the underlying articulated 3D representation.
In particular, MagicPony represents the appearance of an object using a single feature vector and is unable to model high-fidelity details in the input image.
Following~\cite{wu2023magicpony}, for the visualizations with texture rendering, we fine-tune the appearance network for 100 iterations on each test image to obtain sharper textures.
This step is optional but results in a more faithful texture at a modest additional cost (the 3D shape prediction takes 0.5 seconds and the optional texture finetuning requires up to 30 seconds, vs.~hours of distillation approaches like DreamFusion).
Notably, these textures still generalize to the unobserved parts of the object owing to the underlying symmetric neural field representation.

\subsection{Segmentation Masks}
After generating the images $I$, we follow \magicpony and obtain masks using PointRend~\cite{kirillov2019pointrend}, which is trained on COCO~\cite{lin2014microsoft}. Unlike \magicpony, instead of focusing on a single category (\ie, horse), our method works on a variety of quadruped animal categories, some of which (\ie, pig) do not exist in COCO. In practice, we find PointRend also produces reasonably good masks for pigs when asked to detect its training categories (\eg, cow, elephant, sheep, or horse), despite the semantic mismatch.
We, therefore, use the mask with the highest confidence score from detections of these categories.

\subsection{Automatic Object Segmentation via Cross-attention}\label{s:automask}
\begin{figure}[t]
    \centering
    \includegraphics[trim={0px 0px 30px 0px}, clip, width=\linewidth]{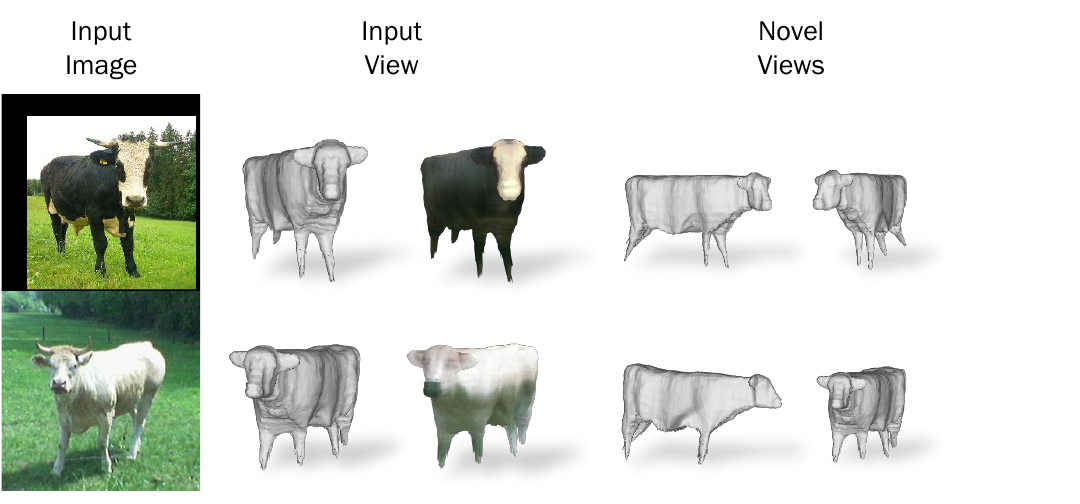}
    \caption{\textbf{Qualitative Results Obtained from Automatic Object Segmentation on Cows.}
    Despite using segmentation masks automatically obtained via cross-attention, our model can still produce 3D shapes that are faithful to the input images and realistic when rendered from novel views.
    }%
    \label{fig:automask}
\end{figure}

We also test our model on unreal ``fictional'' objects, such as steampunk cows, generated by Stable Diffusion.
Since PointRend cannot detect such objects in general,
we adapt a technique based on cross-attention originally developed for interpreting Stable Diffusion~\cite{tang2022daam}, and automatically obtain reasonably accurate masks without manual supervision.
This method produces attribution heatmaps for a given word in the textual prompt used to generate the image.
We condition the generation of the attribution heatmap on the name of the object we are generating (\eg `cow', `horse', `sheep', \etc), and threshold the generated heatmap to obtain an initial segmentation map, which is then refined using the classic GrabCut algorithm~\cite{rother2004grabcut}.
This simple unsupervised segmentation method can also replace PointRend in our training pipeline as demonstrated in~\cref{fig:automask}.

\begin{figure}[!ht]
    \centering
    \includegraphics[trim={0px 5px 0px 0px}, clip, width=\linewidth]{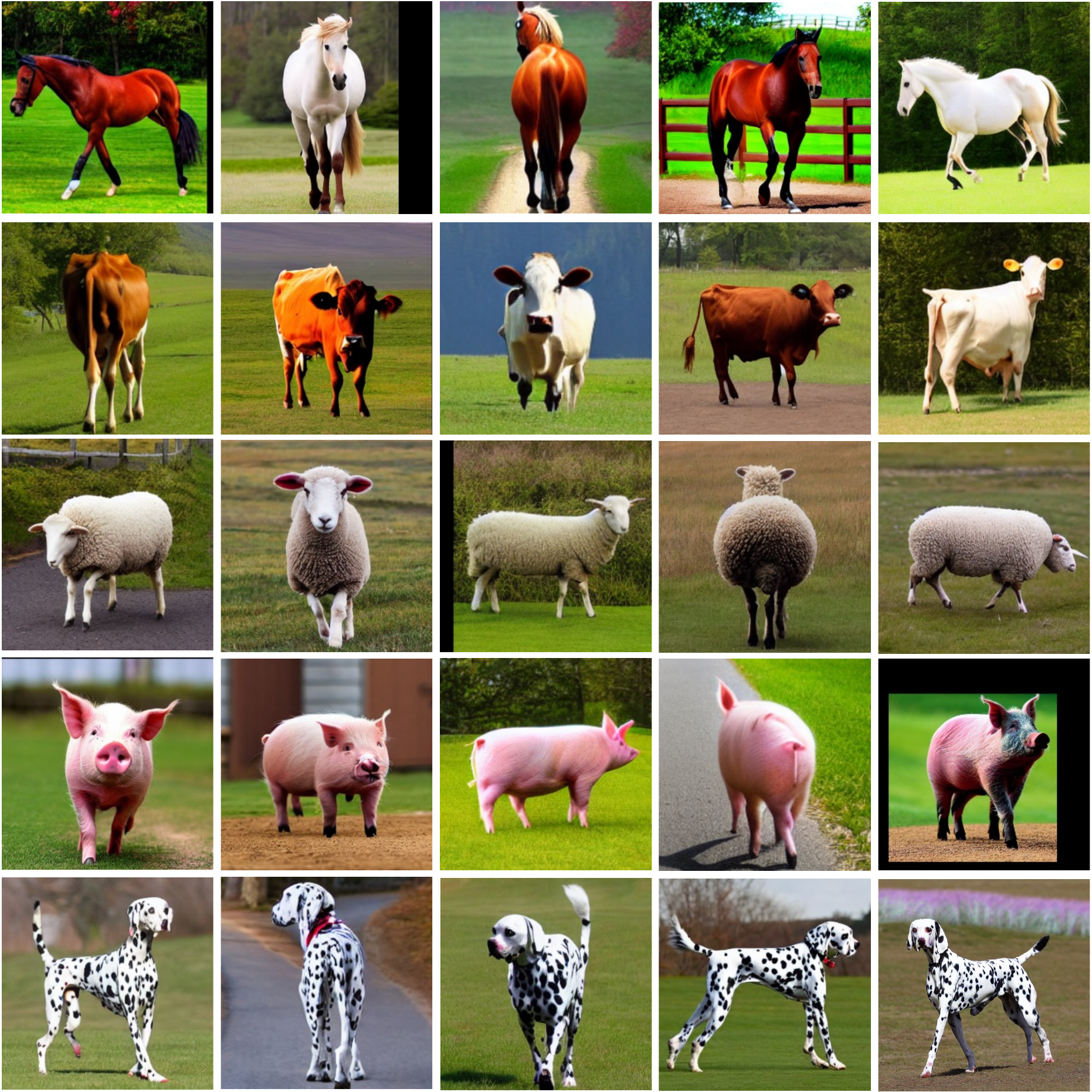}
    \caption{\textbf{Training Images Generated with Stable Diffusion.}
    Additional examples of our generated training images (after cropping).
    }%
    \label{fig:synth-images-sup}
\end{figure}

\begin{table*}[]
\centering
\caption{Effect of prompt engineering (\% of back-views $\uparrow$, \% of unsuitable images $\downarrow$ , PCK@0.1 in keypoint transfer $\uparrow$).}
\begin{tabular}{lllll}
   \toprule
      & horse & cow & sheep   \\ 
   \midrule
   \emph{a \texttt{[V]}} & 3 / 26 / 0.215 & 1 / 29 / 0.185 & 3 / 30 / 0.170  \\
   \emph{a photograph of a single \texttt{[V]}} & 1 / 25 / 0.230 & 1 / 22 / 0.193 & 3 / 5 / 0.183  \\
   \emph{\quad +, \texttt{[V]} is walking away from the camera} (ours) & \textbf{17} / \textbf{4} / \textbf{0.425} & \textbf{10} / \textbf{10} / \textbf{0.402} & \textbf{23} / \textbf{4} / \textbf{0.361} \\
   \bottomrule
\end{tabular}
\label{tab:rebuttal_prompt}
\end{table*}

\subsection{Removal of Truncated Images}
Stable Diffusion sometimes generates portraits or truncated images of an instance, not showing its full body and hence not useful for our purpose of learning a statistical model of an articulated object category.
We filter out such images automatically with a simple heuristic:
if any of the four borders of a generated image, with a 10-pixel margin, contain more than 25\% of masked pixels, we consider the image to be truncated and exclude it from the training dataset.

\subsection{Prompt Design}
\label{s:prompt-design}
To generate training images using Stable Diffusion, we needed to create an appropriate textual prompt.
Starting with a simple prompt, \texttt{a [V]} (where \texttt{[V]} represents the word for an animal category, such as horse, cow, or sheep), we iteratively adapted the prompt until we observed a reasonable quality of generated images, displaying good viewpoint coverage.
This process took about eight iterations.
The final prompt is as follows:
\texttt{a photograph of a single [V], [V] is walking away from the camera}. Additionally, we utilize the following negative prompt:
\texttt{wrong anatomy, animal is lying on the ground, lying, dead, black and white photography, human, person}. \
This single prompt, in conjunction with the negative prompt, generalizes effectively across all categories.
We show examples of generated images in~\cref{fig:synth-images-sup}. Note this engineered prompt is only used for data generation. For SDS gradient computation, we condition Stable Diffusion simply on the prompt \texttt{a [V]} without negative prompts.

We also assess the effectiveness of this prompt design, considering both the percentage of generated back-views of an animal, the percentage of unsuitable images (e.g. cartoons, truncated parts) and the end-to-end performance on the keypoint transfer metric in~\cref{tab:rebuttal_prompt}. 
We start with the initial naive prompt \texttt{[V]} and we gradually increase the complexity by each row.
Each added component improves the quality of generated data.
Most importantly, our final prompt significantly increases the number of back-views. 

\subsection{Implementation Details}
Here we provide additional implementation details not covered in the main text.
We employ the open-source HuggingFace Diffusers library~\cite{von-platen-etal-2022-diffusers} for the Stable Diffusion implementation.
We use Stable Diffusion version 1.5, as we observed that version 2.1 more frequently results in incorrect anatomies.
When generating training images with Stable Diffusion, we set the guidance scale to 10 and perform 50 inference steps using half-precision floating point.
The SDS loss used during training, Eq. (2) in the main text, is weighted with $1\times 10^{-4}$.
The complete training takes 20 hours for 120 iterations on a single NVIDIA A40 GPU.

\section{Datasets}
\subsection{Animodel - 3D Articulated Animals Dataset}

Our dataset consists of two models per category (male and female), with featuring 1-2 textures, created by a professional 3D artist.
We have also collected a set of 28 HDRI backgrounds used to simulate both the background and lighting.
The sources of the random HDRI backgrounds are listed in \cref{tab:hdri}.
We render $1k$ images per category, using a random model from the category and with randomised animation frames, viewpoints, and HDRI backgrounds with randomly sampled azimuth orientation.

\begin{table}[]
\centering
\caption{3D Articulated Animals Dataset HDRI Files URLs}
\resizebox{\linewidth}{!}{
\begin{tabular}{l}
    \url{https://polyhaven.com/a/alps_field} \\
    \url{https://polyhaven.com/a/belfast_farmhouse} \\
    \url{https://polyhaven.com/a/belfast_open_field} \\
    \url{https://polyhaven.com/a/belfast_sunset} \\
    \url{https://polyhaven.com/a/bismarckturm_hillside} \\
    \url{https://polyhaven.com/a/clarens_midday} \\
    \url{https://polyhaven.com/a/drackenstein_quarry} \\
    \url{https://polyhaven.com/a/etzwihl} \\
    \url{https://polyhaven.com/a/evening_meadow} \\
    \url{https://polyhaven.com/a/farm_sunset} \\
    \url{https://polyhaven.com/a/forest_grove} \\
    \url{https://polyhaven.com/a/graveyard_pathways} \\
    \url{https://polyhaven.com/a/green_point_park} \\
    \url{https://polyhaven.com/a/je_gray_02} \\
    \url{https://polyhaven.com/a/je_gray_park} \\
    \url{https://polyhaven.com/a/kloofendal_43d_clear} \\
    \url{https://polyhaven.com/a/lilienstein} \\
    \url{https://polyhaven.com/a/meadow} \\
    \url{https://polyhaven.com/a/meadow_2} \\
    \url{https://polyhaven.com/a/promenade_de_vidy} \\
    \url{https://polyhaven.com/a/resting_place} \\
    \url{https://polyhaven.com/a/resting_place_2} \\
    \url{https://polyhaven.com/a/rosendal_mountain_midmorning} \\
    \url{https://polyhaven.com/a/rosendal_park_sunset} \\
    \url{https://polyhaven.com/a/rural_asphalt_road} \\
    \url{https://polyhaven.com/a/rural_crossroads} \\
    \url{https://polyhaven.com/a/scythian_tombs_2} \\    
\end{tabular}
}   
\label{tab:hdri}
\end{table}

\subsection{ImageNet Images}
In order to train MagicPony on data without manual curation, we automatically collect images from ImageNet~\cite{imagenet} for synsets corresponding to our categories.
We then apply the same pre-processing procedure as for our Stable Diffusion generated data to obtain segmentation masks and remove truncated objects.
This method allows us to obtain $18.4k$, $6.6k$, and $7.7k$ images for horse, cow, and sheep respectively.

\end{document}